\documentclass[pdflatex,sn-mathphys-num]{sn-jnl}
\usepackage{lmodern} 
\usepackage[T1]{fontenc} 
\usepackage{chapterbib}
\usepackage{graphicx}%
\usepackage{titlesec}
\usepackage{makecell}
\usepackage{multirow}
\usepackage{booktabs}
\usepackage{amsmath,amssymb,amsfonts}%
\usepackage{amsthm}%
\usepackage{mathrsfs}%
\usepackage[title]{appendix}%
\usepackage{xcolor}%
\usepackage{url}
\usepackage{textcomp}%
\usepackage{subfig}
\usepackage{manyfoot}%
\usepackage[compatibility=false]{caption}
\usepackage{hyperref}
\hypersetup{
    linktoc=all,
    colorlinks=true,
    linkcolor=blue,
    citecolor=blue,
    urlcolor=blue
}
\usepackage{algorithm}%
\usepackage{algorithmicx}%
\usepackage{algpseudocode}%
\usepackage{listings}%
\usepackage{float}%
\usepackage{tabularx} 
\usepackage{anyfontsize}
\usepackage{tikz}
\usetikzlibrary{positioning}
\titleclass{\subsubsubsection}{straight}[\subsubsection]
\newcounter{subsubsubsection}[subsubsection]
\renewcommand\thesubsubsubsection{\thesubsubsection.\arabic{subsubsubsection}}
\titleformat{\subsubsubsection}
{\normalfont\normalsize\bfseries}{\thesubsubsubsection}{1em}{}
\titlespacing*{\subsubsubsection}
{0pt}{3.25ex plus 1ex minus .2ex}{1.5ex plus .2ex}

\begin{document}

\title[]{Hyperparameter Optimization of Constraint Programming Solvers}


\author*[1]{\fnm{Hedieh} \sur{Haddad}}\email{hedieh.haddad@uni.lu}
\author[2]{\fnm{Thibault} \sur{Falque}}\email{thibault.falque@uni.lu}
\author[2]{\fnm{Pierre} \sur{Talbot}}\email{pierre.talbot@uni.lu}
\author[2]{\fnm{Pascal} \sur{Bouvry}}\email{pascal.bouvry@uni.lu}

\affil*[1]{\orgdiv{Interdisciplinary Centre for Security, Reliability and Trust (SnT)}, \orgname{University of Luxembourg}, \orgaddress{\state{Esch-sur-Alzette}, \country{Luxembourg}}}

\affil[2]{\orgdiv{Faculty of Science, Technology and Medicine (FSTM)}, \orgname{University of Luxembourg}, \orgaddress{\state{Esch-sur-Alzette}, \country{Luxembourg}}}


\abstract{
The performance of constraint programming solvers is highly sensitive to the choice of their hyperparameters. Manually finding the best solver configuration is a difficult, time-consuming task that typically requires expert knowledge. In this paper, we introduce probe and solve algorithm, a novel two-phase framework for automated hyperparameter optimization integrated into the CPMpy library. This approach partitions the available time budget into two phases: a probing phase that explores different sets of hyperparameters using configurable hyperparameter optimization methods, followed by a solving phase where the best configuration found is used to tackle the problem within the remaining time.

We implement and compare two hyperparameter optimization methods within the probe and solve algorithm: Bayesian optimization and Hamming distance search. We evaluate the algorithm on two different constraint programming solvers, ACE and Choco, across 114 combinatorial problem instances, comparing their performance against the solver's default configurations.

Results show that using Bayesian optimization, the algorithm outperforms the solver's default configurations, improving solution quality for ACE in 25.4\% of instances and matching the default performance in 57.9\%, and for Choco, achieving superior results in 38.6\% of instances. It also consistently surpasses Hamming distance search within the same framework, confirming the advantage of model-based exploration over simple local search. Overall, the probe and solve algorithm offers a practical, resource-aware approach for tuning constraint solvers that yields robust improvements across diverse problem types.
}




\maketitle

\section{Introduction}
\label{sec1}

In the field of combinatorial optimization, constraint programming (CP) stands out as a powerful paradigm for solving a wide range of complex problems. It finds applications in diverse domains including scheduling and resource 
management~\citep{baptiste_constraint-based_2001}, network design, 
and logistics~\citep{rossi_handbook_2006}. The core principle of CP involves modeling real-world problems as a set of variables, each with a specific domain of possible values, and a set of constraints that must be satisfied~\citep{rossi_handbook_2006}. A CP solver then systematically searches for one or more solutions that adhere to all specified constraints.

While CP offers remarkable flexibility, the performance of a solver is heavily dependent on its hyperparameter tuning. 
The key hyperparameters include propagation levels~\citep{bessiere_chapter_2006}, which determine the extent of constraint filtering and inference; and the search strategy~\citep{pesant_view_1996}, which dictates how the solver explores the solution space. The search strategy is primarily defined by the choice of variable and value selection heuristics, often referred to as branching rules~\citep{marty_learning_2024}. In addition, the management of computational resources, such as time and memory limits ~\citep{amini_learning_2023,osullivan_opportunities_2021}, while distinct from algorithmic hyperparameters, plays a critical role in ensuring practical solver performance.

These hyperparameters dictate how the solver explores the search space and manages computational resources. An appropriate choice of hyperparameters can lead to dramatic improvements in solver efficiency and solution quality.

However, manually identifying an optimal set of hyperparameters is a daunting task. The search space of possible hyperparameters is vast, and the ideal configuration often varies significantly from one problem instance to another. This manual tuning process is not only time-consuming but also prone to human error~\citep{bischl_hyperparameter_2021}. 

The general problem of finding a set of hyperparameters that improves solver performance, known as algorithm configuration, has a long history in solving hard combinatorial problems~\citep{bischl_hyperparameter_2021}. In recent years, this field has seen rapid progress under the banner of hyperparameter optimization (HPO), largely driven by its widespread adoption and success in machine learning (ML)~\citep{wu_hyperparameter_2019}. Surprisingly, its application within the CP domain remains comparatively limited. The non-trivial task of adapting HPO techniques to the discrete and highly structured search spaces of CP solvers presents a unique research opportunity. These limitations give rise to three main challenges: (i) the combinatorial explosion of possible hyperparameters, (ii) the lack of generalization across problem instances, and (iii) the need for expert knowledge to achieve good performance~\citep{hutter_automated_2019}.

For instance, in the ACE solver~\citep{lecoutre_ace_2024}, the possible parameter combinations exceed 650 million, making exhaustive methods such as grid search computationally infeasible. 
This complexity underscores the necessity of automated hyperparameter optimization methods capable of efficiently exploring such vast hyperparameter spaces.

To address these challenges, we introduce a framework named the probe and solve algorithm (PSA). The core intuition behind PSA is to automate the search for an optimal or near-optimal set of hyperparameters by splitting a given time budget into two distinct phases. First, a probing phase quickly explores a diverse set of hyperparameters by running the solver for short durations on the specific problem instance, gathering valuable performance data about which strategies work well. Second, the solving phase commits the entire remaining time budget to running the solver with the single best-performing configuration identified during probing. This two-phase approach intelligently manages the trade-off between exploring for good configurations and exploiting the best one found.

Basic HPO methods, such as grid search and random search, offer simple starting points but come with notable drawbacks~\citep{bischl_hyperparameter_2021}. Grid search exhaustively evaluates a predefined set of hyperparameters, making it computationally intractable as the number of parameters grows~\citep{liashchynskyi_grid_2019}. Random search provides better efficiency but lacks a guided strategy, often failing to concentrate on the most promising regions of the search space~\citep{bergstra_random_2012}.

A more sophisticated and efficient alternative is Bayesian optimization (BO), a model-based approach that intelligently navigates the hyperparameter space~\citep{ungredda_bayesian_2021}. BO works by building a probabilistic surrogate model (e.g., a Gaussian process~\citep{rasmussen_gaussian_2004}) of the objective function and uses an acquisition function to balance the trade-off between exploration (sampling uncertain, promising new configurations) and exploitation (refining known good configurations)~\citep{candelieri_mastering_2023}. This makes BO particularly well-suited for optimizing expensive black-box functions, such as the performance of a CP solver.

In this paper, we introduce a novel framework for the automated configuration of constraint solvers that bridges the gap between modern, ML-driven HPO techniques and the structured, combinatorial nature of CP.

The contributions of this paper are as follows:

\begin{itemize}
    \item 
    We propose PSA, a resource-aware, two-phase framework for HPO of constraint solvers under fixed time budgets (Section~\ref{sec:psa}).
    
    \item 
    We provide the first integration of the ACE solver into the CPMpy modeling library, enabling automated configuration of its 150,000+ parameter space (Section~\ref{sec:experimentalsetup}).
    
    \item 
    We conduct a large-scale empirical study comparing PSA with BO and Hamming distance search across 114 XCSP3 benchmark instances using two solvers (ACE and Choco) (Section~\ref{sec:experimentalsetup}).
    
    \item 
    We demonstrate that BO within PSA consistently outperforms both default configurations and Hamming distance search, improving solutions for ACE in 25.4\% of instances and Choco in 38.6\% (Section~\ref{sec:results}).

\end{itemize}

These findings offer a deeper understanding of how to steer adaptive solvers and can inform future research in automated solver design.

The remainder of this paper is organized as follows.
Section \ref{sec:background} provides background on constraint programming and hyperparameter optimization methods.
Section \ref{sec:psa} presents PSA and describes its modular architecture and time management strategies.
Section \ref{sec:experimentalsetup} outlines the experimental setup, including benchmark instances, solver configurations, and data collection procedures.
Section \ref{sec:results} reports the main empirical results, comparing against the baselines, and analyzing their components.
Section \ref{sec:discussion} interprets these findings and discusses their implications for automated solver configuration.
Finally, section \ref{sec:conclusion} concludes the paper and highlights avenues for future research.

\section{Background}
\label{sec:background}

\subsection{Constraint Programming}

CP is a declarative paradigm for solving complex combinatorial problems. Its core principle is the separation of problem modeling from the solving algorithm, allowing users to state the problem's logic without specifying the exact steps to find a solution~\citep{lecoutre_constraint_2009}.

At its foundation, a CP model is defined as a constraint satisfaction problem (CSP)~\citep{apt_principles_2003}. A CSP is a triplet \((\mathcal{X}, \mathcal{D}, \mathcal{C})\), where:
\begin{itemize}
    \item \(\mathcal{X} = \{x_1, \ldots, x_n\}\) is a finite set of variables.
    \item \(\mathcal{D} = \{D(x_1), \ldots, D(x_n)\}\) is a set of domains, where each \( D(x_i) \subseteq \mathbb{Z} \) is the finite set of possible values for variable \(x_i\).
    \item \(\mathcal{C} = \{C_1, \ldots, C_m\}\) is a set of constraints that restrict the allowed combinations of values for variables in their scope.
\end{itemize}

An assignment is a mapping that associates each variable \(x_i \in \mathcal{X}\) with one value from its domain \(D(x_i)\).
A solution to a CSP is an assignment that satisfies all constraints in \(\mathcal{C}\).
Let \(\mathcal{A}\) denote the set of all possible assignments:
\[
\mathcal{A} = \{\,A \mid A : \mathcal{X} \rightarrow \bigcup_{x_i \in \mathcal{X}} D(x_i),\; A(x_i) \in D(x_i)\; \text{for all}\; x_i \in \mathcal{X}\,\}.
\]

Many real-world problems require finding not just \textit{any} solution, but the \textit{best} one according to some criterion. This extends the CSP into a constraint optimization problem (COP). A COP is formally defined as a quadruplet  \((\mathcal{X}, \mathcal{D}, \mathcal{C}, f)\), where \((\mathcal{X}, \mathcal{D}, \mathcal{C})\) is a CSP and \(f\) is an \textit{objective function}. This function maps every assignment of variables to an integer value. Formally, if $\mathcal{A}$ is the set of all possible assignments, the signature is $f: \mathcal{A} \rightarrow \mathbb{Z}$.
The goal of a COP solver is to find a solution that satisfies all the constraints of \(\mathcal{C}\) while minimizing (or maximizing) the value of \(f\)~\citep{lecoutre_constraint_2009,hooker_constraint_2018}.


\paragraph{Example: Bounded Knapsack Problem}
A classic example of a COP is the bounded knapsack problem~\citep{cacchiani_knapsack_2022}. Given a set of items, each with a weight and a value, the goal is to determine the number of each item to include in a collection so that the total weight is less than or equal to a given limit and the total value is as large as possible. A CP model for this can be formulated as follows:
\begin{itemize}
    \item \textbf{Variables:} A set of N integer variables, $\mathcal{X} = \{x_1, x_2, \ldots, x_N\}$, where $x_i$ represents the number of times item $i$ is taken.
    \item \textbf{Domains:} Each variable $x_i$ has the domain $D(x_i) = \{0, 1, \ldots, k_i\}$, where $k_i$ is an upper bound on the quantity of item $i$.
    \item \textbf{Constraint:} The sum of the weights of the chosen items must not exceed the knapsack's capacity, $W_{max}$:
    \[
    \sum_{i=1}^{N} w_i \cdot x_i \leq W_{max}
    \]
    \item \textbf{Objective function:} The goal is to maximize the total value of the items in the knapsack:
    \[
    \text{maximize} \quad f(\mathcal{X}) = \sum_{i=1}^{N} v_i \cdot x_i
    \]
\end{itemize}
A solver's task is to find an assignment for $(x_1, \ldots, x_N)$ that satisfies the weight constraint and maximizes the objective function.

\subsection{Hyperparameter Optimization Methods}
\label{subsec:hpo_methods}

HPO is the process of automating the selection of an optimal set of hyperparameters for a given learning algorithm or solver. Formally, we consider an algorithm $\mathcal{S}$ (CP solver) and a hyperparameters space $\Lambda$. 

In general, HPO requires defining an objective (or loss) function $L$ that quantifies the performance of the solver for a given set of hyperparameters $\lambda \in \Lambda$. This performance metric is typically empirical, calculated by running the solver $\mathcal{S}$ with configuration $\lambda$ on a set of problem instances and measuring, for example, the mean runtime or the quality of the solution found~\citep{bischl_hyperparameter_2021}. This process results in a performance score, which is a real value.

The objective function for HPO is therefore a black-box function $L: \Lambda \rightarrow \mathbb{R}$~\citep{feurer_hyperparameter_2019}. The goal of HPO is to find the configuration $\lambda^*$ that minimizes this function:
\[
\lambda^* = \arg \min_{\lambda \in \Lambda} L(\lambda)
\]

It is crucial to distinguish between the COP objective function $f(a)$, which measures the quality of a single problem solution, and the HPO objective function $L(\lambda)$, which measures the performance of a solver configuration.

\paragraph{Example: Applying HPO to the Knapsack Problem}
Continuing with the bounded knapsack example, a CP solver must decide which variable to branch on next (e.g., the item with the best value-to-weight ratio) and which value to try first (e.g., the maximum possible quantity). These choices are key hyperparameters that define the search strategy.
\begin{itemize}
    \item \textbf{Hyperparameter space $\Lambda$}: The space of all possible search strategies. For instance, a single configuration $\lambda_1 \in \Lambda$ could be a "greedy" strategy:
    \[
    \lambda_1 = \{\text{var\_select: \texttt{'max\_value\_ratio'}, val\_select: \texttt{'indomain\_max'}}\}
    \]
    Another configuration, $\lambda_2 \in \Lambda$, representing a more conservative strategy, could be:
    \[
    \lambda_2 = \{\text{var\_select: \texttt{'first\_fail'}, val\_select: \texttt{'indomain\_min'}}\}
    \]
    \item \textbf{HPO objective function $L(\lambda)$}: To evaluate these strategies on a large knapsack instance, we define $L(\lambda)$ as the best objective value found by the solver within a fixed time limit when using the strategy defined by $\lambda$.
    \item \textbf{HPO goal}: The HPO process then automatically tests different configurations like $\lambda_1$, $\lambda_2$, and many others to find the one, $\lambda^*$, that finds the better total value for the knapsack within the given time.
\end{itemize}

\subsubsection{Grid Search}
Grid search is a simple yet exhaustive method that evaluates all possible combinations of hyperparameter values within predefined ranges. 
Although computationally expensive, it provides a useful baseline because it systematically explores the entire hyperparameter space~\citep{liashchynskyi_grid_2019}.

Let $\Lambda$ denote the set of hyperparameters space, expressed as the Cartesian product of the individual hyperparameter domains:
\[
\Lambda = \mathcal{H}_1 \times \mathcal{H}_2 \times \cdots \times \mathcal{H}_k,
\]
where each $\mathcal{H}_i$ represents the discrete set of candidate values for the $i$-th hyperparameter. 
A configuration $\lambda = (\lambda_1, \ldots, \lambda_k)$ corresponds to one choice of values, where $\lambda_i \in \mathcal{H}_i$. 
The total number of configurations is therefore:
\[
N = \prod_{i=1}^{k} |\mathcal{H}_i|.
\]

The objective function $L(\lambda)$ is evaluated for each configuration $\lambda \in \Lambda$, and the configuration that achieves the lowest value is selected as the best one:
\[
\lambda^* = \arg\min_{\lambda \in \Lambda} L(\lambda).
\]

While grid search guarantees complete coverage of the hyperparameter space, its computational cost grows exponentially with the number of hyperparameters~\citep{bischl_hyperparameter_2021}. 
For example, if a solver has three hyperparameters and each has ten possible values, the total number of evaluations required is $10^3 = 1{,}000$.

\subsubsection{Iterative Search Methods: Random and Local Search}
\label{subsec:iterative_search}

Moving beyond exhaustive methods like grid search, iterative strategies explore the hyperparameter space sequentially. Two fundamental and contrasting approaches are random search and local search, which represent the core principles of global exploration and local exploitation, respectively.

Random search is an uninformed method that focuses purely on exploration. It samples hyperparameters \(\lambda\) at random from the hyperparameter space \(\Lambda\) until a predefined budget is met. The key advantage is its effectiveness in high-dimensional spaces where only a few hyperparameters are critical. Unlike grid search, which systematically evaluates all combinations from predefined discrete grids, random search explores the space more broadly with fewer evaluations, increasing the likelihood of finding near-optimal configurations~\citep{bergstra_random_2012}.
However, its major drawback is that it does not learn from past evaluations; every trial is independent, which can lead to inefficiently resampling of unpromising regions~\citep{bergstra_algorithms_2011}.

In direct contrast, local search is a method centered on exploitation. It begins with an initial configuration and iteratively moves to an adjacent or neighboring configuration only if it offers improved performance. The concept of a neighborhood is crucial, especially in spaces with categorical hyperparameters~\citep{hutter_sequential_2011}. This raises the challenge of how to define a neighbor in a discrete space; the standard approach is to use a distance metric, with the Hamming distance being one of the most fundamental metrics. This is because it is particularly well-suited for categorical data, which has no inherent ordering, by simply counting the number of differing hyperparameter values between two configurations~\citep{hutter_sequential_2011}. Its simplicity and efficient computation make it a valuable and straightforward method for defining a neighborhood in these discrete spaces.

\subsubsection{Hamming Distance}
\label{par:Hamming}
The Hamming distance is a standard metric that measures dissimilarity by counting the positions at which two configuration vectors differ:


\[
d_H(A, B) = \sum_{i=1}^{n} \mathbb{I}(A_i \neq B_i)
\]
where \(\mathbb{I}(\cdot)\) is the indicator function, which evaluates to 1 if its argument is true and 0 otherwise. The neighborhood of a configuration $\lambda$ in $\Lambda$, denoted $\mathcal{N}(\lambda)$, is defined as:
\[
\mathcal{N}(\lambda) = \{\lambda' \in \Lambda \mid d_H(\lambda, \lambda') = 1\}.
\]

These methods exemplify the classic exploration–exploitation trade-off. Random search wastes evaluations in unpromising regions (inefficient exploitation), whereas pure local search can easily become trapped in local optima (poor exploration). 

These challenges have led to hybrid optimization strategies. For instance, the iterated local search (ILS) metaheuristic~\citep{hutter_automatic_2007} is a simple but powerful method. ILS works by repeatedly applying a local search to a solution, then changing it (perturbing) to escape local optima. By switching between refining a solution locally and making random changes, ILS can effectively explore new and promising areas.

A good example of this is ParamILS, an algorithm that uses the ILS approach~\citep{hutter_paramils_2009}. ParamILS searches locally for the best solution, and when it gets stuck, it makes a random change to start searching in a new area. While these iterative methods are better than simple techniques like grid search, they have limits. This shows we need smarter, model-based approaches that intelligently balance exploration and exploitation.

\subsubsection{Bayesian Optimization}
\label{subsubsec:bo}

BO is a powerful, model-based HPO strategy designed to optimize expensive black-box functions. This makes it particularly well suited for tuning computationally costly CP solvers, where each evaluation of a set of hyperparameters can take a significant amount of time.
Unlike uninformed methods such as grid search or random search, BO builds a probabilistic surrogate model to approximate the objective function. 

The method builds a surrogate model \(g(x)\) of the true objective function \(L(x)\), and selects the next configuration \(x_{n+1}\) to evaluate by optimizing an acquisition function \(a(x)\). The role of the acquisition function is to balance the trade-off between exploring uncertain regions of the hyperparameter space and exploiting regions known to have good performance~\citep{candelieri_mastering_2023}. The most common choice for this surrogate is a Gaussian process (GP)~\citep{rasmussen_gaussian_2004}, which defines a prior over functions. After observing some data, this is updated to a posterior distribution that models our belief about the objective function's behavior, providing a mean prediction and an uncertainty estimate for any given configuration:
\[
    g(x) \sim \mathcal{GP}\big(\mu(x), k(x, x')\big)
\]
where:
    \begin{itemize}
        \item \(\mu(x)\) is the mean function, representing the expected value of \(L(x)\).
        \item \(k(x, x')\) is the covariance or kernel function, which measures the similarity between points \(x\) and \(x'\). It models the correlation between the function values at those points.
    \end{itemize}

This surrogate model is then used by an acquisition function to intelligently guide the search for the next configuration to evaluate~\citep{kushner_new_1964,gan_acquisition_2021}. The role of the acquisition function is to balance the critical exploration (probing regions where the model is highly uncertain, which could potentially hide an even better, undiscovered optimum) and exploitation (focusing on regions that the surrogate model predicts will yield high performance) trade-off~\citep{gan_acquisition_2021}.

Standard acquisition functions like expected improvement quantify this potential and select the configuration that offers the best balance~\citep{candelieri_mastering_2023}.

The BO process is iterative: after each new configuration is evaluated by running the CP solver, the result is used to update the surrogate model~\citep{brause_combining_2022}. This allows the search to become progressively more informed, concentrating its evaluations in the most promising areas of the hyperparameter space. By building this explicit model, BO aims to find high-quality solutions with significantly fewer evaluations, a crucial advantage in the CP domain~\citep{brause_combining_2022,amadini_portfolio_2016}.

\section{Probe and Solve Algorithm}
\label{sec:psa}

To systematically optimize a set of hyperparameters for CP solvers, we introduce PSA, a flexible and adaptive framework designed to make effective use of a limited time budget. The core idea of PSA is to partition a global time-limit ($T_g$) into two distinct phases:
\begin{enumerate}
    \item A \textit{probing phase}, where a dedicated portion of the time budget is used to explore a wide range of hyperparameters. This is achieved by running many short-lived, time-limited solver runs, each with a different configuration, to gather performance data.
    \item A \textit{final solving phase}, where the single best-performing configuration identified during probing is used to solve the problem instance with the entire remaining time budget.
\end{enumerate}

This two-phase approach provides a structured balance between the exploration of the hyperparameter space and the exploitation of the most promising strategy found. The framework is designed to be highly modular, allowing different strategies for time management, hyperparameter selection, and timeout evolution to be composed, enabling a thorough investigation of their combined effects.

The PSA framework follows a two-phase architecture that partitions the global time budget $T_g$ into probing time $t_p$ and solving time $t_s$:
\[
t_p = \rho \cdot T_g, \quad t_s = T_g - t_p, \quad \rho \in [0,1]
\]

With $\rho=0.2$ as the default value based on empirical analysis, 20\% of the total time is dedicated to exploring strategies, while 80\% is reserved for actual solving with the best strategy found. This balanced allocation addresses the fundamental trade-off between exploration and exploitation in algorithm configuration~\citep{haddad_selecting_2024}.

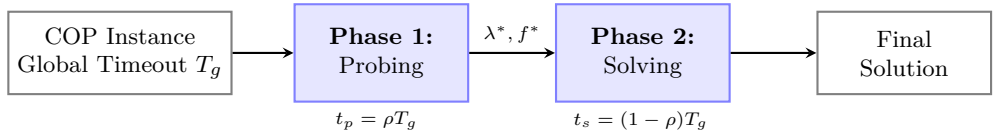
\begin{figure}[h!]
\centering
\resizebox{\linewidth}{!}{%
\begin{tikzpicture}[
    phase/.style={rectangle, draw=blue!50, fill=blue!10, thick,
                  minimum width=2.3cm, minimum height=1.3cm,
                  align=center, font=\small},
    component/.style={rectangle, draw=black!50, fill=white, thick,
                      minimum width=2.3cm, minimum height=1.1cm,
                      align=center, font=\small},
    arrow/.style={->, >=stealth, thick},
    label/.style={font=\footnotesize}
]

\node (input) [component] at (0,0) {COP Instance \\ Global Timeout $T_g$};
\node (phase1) [phase, right=2.3cm of input.center] {\textbf{Phase 1:} \\ Probing};
\node (phase2) [phase, right=2.3cm of phase1.center] {\textbf{Phase 2:} \\ Solving};
\node (output) [component, right=2.3cm of phase2.center] {Final \\ Solution};

\draw [arrow] (input) -- (phase1);
\draw [arrow] (phase1) -- node[above, label] {$\lambda^*, f^*$} (phase2);
\draw [arrow] (phase2) -- (output);

\node [font=\footnotesize] at (3.4,-0.9) {$t_p = \rho T_g$};
\node [font=\footnotesize] at (6.9,-0.9) {$t_s = (1-\rho)T_g$};

\end{tikzpicture}
}
\caption{PSA two-phase architecture with time allocation. The framework partitions the global time budget $T_g$ into probing time $t_p$ and solving time $t_s$.}
\label{fig:psa_architecture}
\end{figure}

Figure~\ref{fig:psa_architecture} illustrates the PSA architecture. The framework begins with a COP instance and global time budget, proceeds through the probing phase to identify the best configuration $\lambda^*$, and concludes with the solving phase using the remaining time.

\subsection{Probing Phase}
\label{subsec:probing_phase}

The goal of the probing phase is to efficiently sample the hyperparameter space to identify a high-quality configuration. The phase consists of a sequence of iterative trials, hereafter referred to as \textit{rounds}. In each round, a single set of hyperparameters is selected and evaluated by running the solver for a short duration. The execution of the probing phase is detailed in Algorithm~\ref{alg:probe_phase} and is controlled by these key orthogonal strategy types.

\begin{algorithm}[htbp] 
\caption{Probing Phase of PSA}
\label{alg:probe_phase}
\begin{algorithmic}[1]
\State \textbf{Input:} COP \((\mathcal{X}, \mathcal{D}, \mathcal{C}, f)\); HPO \(\mathcal{H}\); \(\Lambda\); \(T_g\); \(\tau_0\); \(\mathit{TimeInit}\); \(\mathit{Evolve}\); \(\mathit{stop\_type}\); \(L\).
\State \textbf{Output:} \(\lambda^\star\), \(f^\star\), \(t_{\text{rem}}\).
\Statex
\State \(t_p, t_s \gets \text{AllocateTime}(T_g)\) \Comment{Probe/solve budgets}
\State Initialize: \(f^\star \gets \infty\), \(\lambda^\star \gets \text{default}\), \(\mathit{stag} \gets 0\), \(t_0 \gets \text{Now}()\)
\Statex
\If{\(\mathit{TimeInit} = \textsc{FirstRuntime}\)} \Comment{Adaptive timeout}
    \State \((sol, rt) \gets \text{Solve}(\text{COP}, \text{default}, t_p)\)
    \State \(\tau \gets rt\) \Comment{Use runtime}
    \If{\(sol\)} \(f^\star \gets f(sol)\); \(\lambda^\star \gets \text{default}\) \EndIf
\Else \Comment{Static timeout}
    \State \(\tau \gets \tau_0\)
\EndIf
\Statex
\While{\(\text{Now}() - t_0 < t_p\)} \Comment{Main loop}
    \State \(\lambda \gets \mathcal{H}.\text{NextConfig}()\)
    \If{\(\lambda = \varnothing\)} \textbf{break} \EndIf
    \Statex
    \If{$\lambda \in \mathcal{M}$} \Comment{Check memory array}
        \State $(obj, rt) \gets \mathcal{M}[\lambda]$ \Comment{Retrieve cached result}
    \Else
        \State $(sol, rt) \gets \text{Solve}(\text{COP}, \lambda, \tau)$
        \State $obj \gets f(sol)$ if $sol$ else $\infty$
        \State $\mathcal{M}[\lambda] \gets (obj, rt)$ \Comment{Store in memory array}
    \EndIf
    \Statex
    \If{\(obj < f^\star\)} \Comment{Better solution found}
        \State \(f^\star \gets obj\); \(\lambda^\star \gets \lambda\); \(\mathit{stag} \gets 0\)
    \Else
        \State \(\mathit{stag} \gets \mathit{stag} + 1\)
    \EndIf
    \Statex
    \State \(\mathcal{H}.\text{UpdateModel}(\lambda, obj, rt)\)
    \State \(\tau \gets \mathit{Evolve}(\tau)\) \Comment{Update timeout}
    \Statex
    \If{\(\mathit{stop\_type} = \textsc{FirstSolution}\) and \(obj \neq \infty\)} \textbf{break} \EndIf
    \If{\(\mathit{stop\_type} = \textsc{Stagnation}\) and \(\mathit{stag} \ge L\)} \textbf{break} \EndIf
\EndWhile
\Statex
\State \(t_{\text{rem}} \gets T_g - (\text{Now}() - t_0)\)
\State \textbf{return} \((\lambda^\star, f^\star, t_{\text{rem}})\)
\end{algorithmic}
\end{algorithm}

\subsubsection{HPO Method}
This is the core HPO engine of the PSA, responsible for selecting the next set of hyperparameters to evaluate. To ensure modularity, the framework requires any HPO strategy to implement a specific interface with three key functions:
\begin{enumerate}
    \item \textbf{\texttt{NextConfig()}:} Returns the next set of hyperparameters to be evaluated from the search space.
    \item \textbf{\texttt{UpdateModel(params, runtime)}:} Takes the parameters of the last trial and its resulting runtime and objective to update the strategy's internal model. For model-free methods like Hamming distance, this may simply involve updating the incumbent.
    \item \textbf{\texttt{AllocateTime($T_g$)}:} Partitions the global timeout into probing and solving budgets:
      \[
      \texttt{AllocateTime}(T_g) = (\rho T_g,\ (1-\rho)T_g)
      \]
      where $\rho$ is the probing ratio (default $\rho=0.2$). This creates a clear separation: $t_p$ for configuration exploration, $t_s$ for solving with the best configuration.
    \item \textbf{\texttt{GetBestConfig()}:} Returns the best configuration found so far, which is used in the final solving phase.
\end{enumerate}
We implement BO as a core component of PSA. It is a concrete strategy that conforms to this interface. BO is a sophisticated model-based strategy where \texttt{NextConfig} uses an acquisition function over a GP model to choose the next point. \texttt{UpdateModel} retrains the GP with the new data point.

\subsubsection{Global Time Management}
This high-level component determines the total time budget allocated to the probing phase. We implement two approaches:
\begin{itemize}
    \item \textbf{Static percentage allocation:} A fixed percentage of $T_g$ is reserved for the entire probing phase. For example, with a $T_g$ of 1800 seconds and a 20\% allocation, the probing phase will run for a total of 360 seconds before automatically transitioning to the solving phase. This ensures a predictable and consistent time split across different experiments.
    \item \textbf{Iteration-limited allocation:} The probing phase continues until either the $T_g$ is exhausted or a predefined maximum number of configuration trials (\texttt{max\_tries}) has been completed. This strategy allows for more flexible exploration, as the total probing time is determined by the cumulative runtime of the trials, making it suitable for scenarios where individual probe runtimes are highly variable.
\end{itemize}

\subsubsection{Round Timeout Initialization}
This component determines the timeout for each individual configuration evaluation within the probing phase. This sets the time limit for the first configuration to be tested.
\begin{itemize}
    \item \textbf{Static initial timeout:} A fixed, small timeout (e.g., 5 seconds) is used. This provides a consistent baseline but may be too short or too long for certain problem instances.
    \item \textbf{First-runtime timeout:} The framework first performs a preliminary run of the solver using its default configuration. This run is allocated an informed time limit which is up to the total available probe budget, and will stop as soon as the first solution is found. The actual runtime observed from this initial run is then used as the timeout for subsequent probes, thereby adapting the timeout to the intrinsic difficulty of the problem instance.
\end{itemize}

\subsubsection{Timeout Evolution Pattern}
After the initial round, the timeout for subsequent rounds can evolve to adapt to the performance of previously tested configurations. Our framework manages this through a modular \textit{Timeout Evolution Strategy}, which implements an \texttt{Evolve(current\_timeout)} function to compute the timeout for the next round based on the value from the current one.

We implement and test three concrete instances of this strategy:
\begin{itemize}
    \item \textbf{Static evolution:} This strategy's \texttt{Evolve} function simply returns the same \texttt{current\_timeout} it was given, keeping the timeout fixed.
    \item \textbf{Geometric evolution:} Here, the $\texttt{Evolve}(t_i)$ function returns the current timeout multiplied by a predefined growth factor, $\beta > 1$. For example, $t_{i+1} = \texttt{Evolve}(t_i) = t_i \times \beta $.
    \item \textbf{Luby sequence evolution:} This strategy's $\texttt{Evolve}(t_i)$ function maintains an internal counter $k$. It returns $t_{i+1} = t_0 \cdot \text{Luby}(k)$, where $t_0$ is the initial timeout and $\text{Luby}(k)$ is the $k$-th value in the sequence (1, 1, 2, 1, 1, 2, 4, ...), incrementing $k$ after each call.
\end{itemize}

\subsubsection{Probing Stop Conditions}
The probing phase is allocated a specific time budget (e.g., a percentage of the global time limit or up to a maximum number of tries). The phase will terminate when this budget is exhausted, but it can also be configured to stop early to pivot to the final solving phase more quickly. The defined stopping conditions are:
\begin{itemize}
    \item \textbf{Timeout}: The primary stopping condition is the exhaustion of the allocated probing time budget. The phase will always terminate when its time is up, ensuring that the final solving phase receives its planned portion of the global time limit.
    \item \textbf{First solution found:} The probing phase can be configured to halt as soon as any configuration finds the first feasible solution. This is particularly useful in scenarios where quickly finding any valid solution is more critical than extensive exploration for the absolute best configuration.
    \item \textbf{Stagnation:} The process can terminate if the best-found solution has not improved after a predefined number of rounds. This prevents the framework from wasting time on continued exploration if the HPO strategy appears to have converged or is no longer making progress.
\end{itemize}

\subsubsection{General Improvements}
To enhance the efficiency and scalability of our approach, we incorporate a memory-based mechanism:
\begin{itemize}
    \item \textbf{Memory array:} A data structure is implemented to store previously evaluated configurations and their corresponding results. When a configuration is revisited, the solver retrieves the stored results instead of performing redundant evaluations. This not only accelerates the optimization process, but also avoids redundant computation, particularly for large-scale problem instances.
\end{itemize}

\subsection{Solving Phase}
\label{subsec:solving_phase}

Once the probing phase concludes, PSA transitions to the solving phase, which utilizes the best-found configuration to solve the instance with the remaining time budget.

\begin{algorithm}[htbp]
\caption{Solving Phase of PSA}
\label{alg:solve_phase}
\begin{algorithmic}[1]
\State \textbf{Input:} COP $(\mathcal{X}, \mathcal{D}, \mathcal{C}, f)$; HPO $\mathcal{H}$; $\lambda^\star$; $f^\star$; $t_{\text{rem}}$.
\State \textbf{Output:} $sol_{\text{final}}$, $f(sol_{\text{final}})$.
\Statex
\If{$f^\star = \infty$} \Comment{No solution found in probing}
    \State $\lambda_{\text{final}} \gets \mathcal{H}.\text{GetBestConfig}()$
    \State $(sol_{\text{final}}, rt) \gets \text{Solve}(\text{COP}, \lambda_{\text{final}}, t_{\text{rem}})$
\Else \Comment{Incumbent exists; try to improve}
    \State $\mathcal{C}' \gets \mathcal{C} \cup \{f(\mathcal{X}) < f^\star\}$ \Comment{Add cut constraint}
    \State $(sol_{\text{final}}, rt) \gets \text{Solve}((\mathcal{X}, \mathcal{D}, \mathcal{C}', f), \lambda^\star, t_{\text{rem}})$
\EndIf
\Statex
\State \textbf{return} $(sol_{\text{final}}, f(sol_{\text{final}}))$
\end{algorithmic}
\end{algorithm}

The solving phase operates as follows:
\begin{enumerate}
    \item The single best-performing set of hyperparameters (\(\lambda^\star\)) identified during the probing phase is retrieved.
    \item If any solution was found during probing (i.e., \(f^\star < \infty\)), a new objective cut constraint is added to the model: \(\mathcal{C}' \leftarrow \mathcal{C} \cup \{f(\mathcal{X}) < f^\star\}\). This constraint forces the solver to search only for solutions that are strictly better than the current best, implementing an optimization loop that focuses the search on improvement rather than re-finding known solutions.
    \item The solver is executed with \(\lambda^\star\) and the (potentially updated) constraint model \(\mathcal{C}'\), using the entire remaining time budget (\(T_g - \text{probing elapsed time}\)).
\end{enumerate}

\subsection{Complete PSA Algorithm Specification}
\label{subsec:complete_psa_algorithm}

The integrated PSA algorithm, presented in Algorithm~\ref{alg:psa_complete}, combines the probing and solving phases into a unified framework. This algorithm operationalizes the two-phase tuning strategy by integrating the modular components for HPO and time management described previously.

\begin{algorithm}[h!]
\caption{Complete PSA Framework}
\label{alg:psa_complete}
\begin{algorithmic}[1]
\State \textbf{Input:} COP instance \((\mathcal{X}, \mathcal{D}, \mathcal{C}, f)\); global time limit \(T_g\); probing fraction \(\rho\); HPO method \(\mathcal{H}\); hyperparameter space \(\Lambda\).
\State \textbf{Output:} Final solution \(\mathit{sol_{final}}\) and its objective value \(f(\mathit{sol_{final}})\).
\Statex
\State \(t_p \gets \rho \times T_g\) \Comment{Calculate probing budget}
\State \(t_s \gets T_g - t_p\) \Comment{Calculate solving budget}
\Statex
\State \textit{Phase 1: Probing}
\State \((\lambda^\star, f^\star) \gets \text{ProbingPhase}((\mathcal{X}, \mathcal{D}, \mathcal{C}, f), \Lambda, t_p, \mathcal{H})\) \Comment{Execute Algorithm~\ref{alg:probe_phase}}
\Statex
\State \textit{Phase 2: Solving}
\State \((\mathit{sol_{final}}, f_{final}) \gets \text{SolvingPhase}((\mathcal{X}, \mathcal{D}, \mathcal{C}, f), \lambda^\star, f^\star, t_s)\) \Comment{Execute Algorithm~\ref{alg:solve_phase}}
\Statex
\State \textbf{return} \((\mathit{sol_{final}}, f_{final})\)
\end{algorithmic}
\end{algorithm}

This integrated approach provides several key properties:
\begin{itemize}
    \item \textbf{Time-awareness:} The framework strictly respects the global timeout \(T_g\), ensuring practical applicability.
    \item \textbf{Adaptability:} The adaptive timeout mechanism in the probing phase adjusts to problem difficulty.
    \item \textbf{Robustness:} Fallback mechanisms ensure a solution is returned even if individual phases fail.
    \item \textbf{Modularity:} Components (HPO methods, time allocation strategies) can be easily swapped.
\end{itemize}

\section{Experimental Setup}
\label{sec:experimentalsetup}
This section details the environment, problem instances, and the specific configurations employed for our large-scale benchmarking study. We begin by describing the two constraint solvers used as underlying engines, followed by an outline of the problem instances and the comprehensive benchmarking methodology.

To robustly evaluate the PSA framework, we selected two distinct constraint solvers: ACE and Choco. Their varying complexities and feature sets provide a broad and representative testbed for our tuning approach.

\subsection{The ACE Solver}
\label{subsubsec:benchmarks_ace}
ACE is a modern constraint solver, implemented in Java, designed to tackle combinatorial problems involving integer and Boolean variables. Its extensive capabilities comply with the XCSP3-core standard, a widely recognized format for constraint problems~\citep{boussemart_xcsp3-core_2024}. ACE offers a rich library of constraints, including advanced table constraints (ordinary, starred, and hybrid), and a wide array of global constraints such as \texttt{AllDifferent}, \texttt{Cardinality}~\citep{regin_generalized_1996}, \texttt{Count}, \texttt{Element}, and \texttt{Cumulative}~\citep{lecoutre_constraint_2009,bessiere_complexity_2004,baptiste_constraint-based_2001}. Additionally, it features various built-in search heuristics and is optimized for mono-criterion optimization. This comprehensive feature set is particularly valuable for PSA, as it contributes to a vast, complex, and highly influential hyperparameter space, making it an ideal candidate for showcasing the benefits of automated tuning.

The chosen hyperparameter space for ACE, summarized in Table~\ref{tab:benchmarks_ace_hyperparameters}, consists of 9 distinct dimensions that govern the solver's core search behavior. These include heuristics for variable and value selection (\texttt{varh}, \texttt{valh}), the information of table constraints (\texttt{sc2}, \texttt{sc3}), and other categorical parameters that guide branching, learning, and constraint handling. The combination of these chosen options results in a combinatorial search space of over 150 thousand unique configurations. This scale makes exhaustive tuning methods like grid search computationally infeasible and underscores the significant challenge of manual tuning. Consequently, ACE serves as an excellent testbed for demonstrating PSA's ability to effectively navigate such a hyperparameter space.

A practical contribution of this work is the integration of the previously unsupported ACE solver into the CPMpy library~\citep{cpmpy_cpmpy_2025}. Within this newly integrated environment, we apply our HPO strategies to tune ACE across various problem instances. The framework's success is assessed based on both the quality of the final objective value obtained and the total time required to find that solution. Thus, utilizing ACE not only provides a challenging and relevant testbed for our tuning framework but also enhances the capabilities of the broader CPMpy ecosystem for future research.

\begin{table}[!tp]
\caption{The Tunable Hyperparameter Space of the ACE Solver.}
\label{tab:benchmarks_ace_hyperparameters}
\centering
\begin{tabularx}{\textwidth}{@{} l >{\raggedright\arraybackslash}X r @{}}
\toprule
\textbf{Parameter} & \textbf{Description} & \textbf{\# Options} \\
\midrule
\texttt{varh} & Variable Selection Heuristic & 10 \\
\texttt{valh} & Value Selection Heuristic & 10 \\
\texttt{saf}  & Stay Array Focus & 2 \\
\texttt{pc1}  & Preserve Unary Constraints & 2 \\
\texttt{toh}  & Convert to Hybrid Tables & 2 \\
\texttt{lc}   & Last Conflict Weighting & 4 \\
\texttt{negative} & Algorithm for Negative Table Constraints & 3 \\
\texttt{sc2}  & Structure for Binary Table Constraints & 4 \\
\texttt{sc3}  & Structure for Ternary Table Constraints & 4 \\
\midrule
\multicolumn{2}{@{}l}{\textbf{Total Combinatorial Configurations}} & \textbf{153,600} \\
\bottomrule
\end{tabularx}
\end{table}

\subsection{The Choco Solver}
\label{subsubsec:benchmarks_choco}
We also utilize the Choco solver (v4.10.6), a widely recognized open-source CP library implemented in Java. Choco provides a rich collection of heuristics and propagation levels, making it a strong alternative for benchmarking the PSA framework~\citep{prudhomme_choco-solver_2022}.

The hyperparameter space for Choco, which was derived from a JSON file detailing its available hyperparameters, is presented in Table~\ref{tab:benchmarks_choco_tunable_params}. This space includes key components such as restart policies (\texttt{restarts}), variable and value selection heuristics (\texttt{varh}, \texttt{valh}), consistency levels (\texttt{lc}), and propagation queue flushing thresholds (\texttt{flush}).

\begin{table}[!tp]
\centering
\caption{The Tunable Hyperparameter Space of the Choco Solver.}
\label{tab:benchmarks_choco_tunable_params}
\begin{tabular}{l p{0.6\linewidth} r}
\toprule
\textbf{Parameter} & \textbf{Description} & \textbf{\# Options} \\
\midrule
Lc & Level of Consistency (Level of propagations) & 2 \\
Restarts & Restart policy (e.g., LUBY, GEOMETRIC, NONE) & 12 \\
Valh & Value Selection Heuristic & 6 \\
Varh & Variable Selection Heuristic & 19 \\
Flush & Propagation queue flush threshold & 5 \\
\midrule
\multicolumn{2}{l}{\textbf{Total Combinatorial Configurations}} & \textbf{136,800} \\
\bottomrule
\end{tabular}
\end{table}

\subsection{Benchmarking Methodology}
Our benchmarking approach involved four distinct types of experimental runs for both the ACE and Choco solvers:

\paragraph{Default Solver Baselines}
For each of the 114 instances, we first solve the instances with the default version of both ACE and Choco solvers. These runs served as a foundational baseline, reflecting the out-of-the-box performance of each solver without any hyperparameter tuning. Each default run adhered to the same system environment and the 1800-second global time limit.

\paragraph{PSA Framework Configurations}
The core of our extensive benchmarking involves the PSA framework, which employs two distinct HPO methods: BO and Hamming distance search~\citep{schaus_model-based_2022}. Both methods are implemented within the PSA framework, allowing a direct comparison of their effectiveness. We conduct a full-factorial experiment to systematically evaluate the impact of each modular component of the PSA framework. This design allows us to not only identify the best overall configuration but also to analyze the independent contribution of each strategic dimension to performance. A total of 24 unique PSA configurations for each HPO method are generated by combining all possibilities from the following strategic dimensions:
\begin{itemize}
\item \textbf{Probing Time Allocation:} 2 options (\textit{Static}, \textit{Dynamic})
\item \textbf{Round Timeout Initialization:} 2 options (\textit{Static}, \textit{First-Runtime})
\item \textbf{Timeout Evolution Pattern:} 3 options (\textit{Static}, \textit{Geometric}, \textit{Luby})
\item \textbf{Probing Stop Condition:} 3 options (\textit{First Solution}, \textit{Timeout}, \textit{Stagnation})
\end{itemize}

Some combinations of these settings are inherently incompatible. In particular, when the probing stop condition is set to \textit{First Solution}, the probing phase terminates as soon as the first solution is found. In this case, the probing loop executes only a single round, which makes any form of round-timeout evolution irrelevant. For this reason, whenever the stop condition is \textit{First Solution}, only the \textit{static} timeout evolution pattern is meaningful and permitted. Accordingly, these 24 configurations are derived from two batches:
\begin{itemize}
\item Round-timeout static: 2$\times$3$\times$3 = 18 configurations
\item Round-timeout First-Runtime: 2$\times$1$\times$3 = 6 configurations
\end{itemize}
These 24 PSA configurations for each HPO method are evaluated per instance for both the ACE and Choco solvers.

\paragraph{Champion Configuration Evaluation}
Following the exploratory phase of the PSA framework, the single best-performing configuration (the \textit{champion configuration}) for each combination of solver and HPO method is identified based on the aggregated results. Subsequently, these identified champion configurations are then run separately on all 114 instances. These dedicated runs provide a direct and robust measure of their performance for comparative analysis against the baselines.

\subsection{Execution Environment and Data Collection}
To ensure fairness and reproducibility, all experiments were conducted under strictly controlled conditions.
\begin{itemize}
\item \textbf{Global time limit:} Each individual experimental run is allocated a maximum wall-clock time of 1800 seconds (30 minutes), aligning with the standard duration used in the XCSP3 competition from which our benchmarks are drawn.

\item \textbf{Computational resources:} The full suite of experiments was executed on the high-performance computing (HPC) facilities, with the technical specifications of a cluster compute node being 2xAMD Epyc ROME 7H12 @ 2.6 GHz [64c/280W] processor with 256 GB RAM.

\item \textbf{Total computational effort:} The complete experiment, encompassing 6156 individual runs (114 instances $\times$ (24 PSA configurations $\times$ 2 HPO methods + 1 default solver baseline configuration + 1 PSA champion for each HPO method) $\times$ 2 solvers), represents a total of 3078 compute-hours.

\item \textbf{Data logging and reproducibility:} For each run, detailed performance data is logged to structured CSV files:
\begin{itemize}
\item \textbf{Objective value:} The best objective value found by the solver within the time limit.
\item \textbf{Solver status:} The final exit status reported by the solver (e.g., \texttt{OPTIMUM FOUND}, \texttt{SATISFIABLE}, \texttt{TIMEOUT}, \texttt{ERROR}).
\item \textbf{Runtime:} The total wall-clock time elapsed for the run.
\item \textbf{Configuration details:} All command-line flags used for each run are logged to ensure full reproducibility and allow detailed performance analysis.
\end{itemize}


\item \textbf{Parameter Settings for PSA Components:} For strategies utilizing the \textit{Static Percentage Allocation}, the probing budget was set to 20\% of \textit{$T_g$}, a choice that tries to balance exploration with a substantial budget for the final solve. The \textit{Static} initial round timeout was set to a baseline of \textit{5} seconds. For the \textit{Geometric} evolution pattern, a growth factor of $\beta=1.5$ is used. These values are chosen as reasonable, standard defaults to avoid overfitting the framework to this specific benchmark set.
\end{itemize}

\section{Results and Analysis}
\label{sec:results}
This section presents the comprehensive results of our large-scale benchmarking experiments. Our primary objective is to identify the most effective automated strategies for configuring a solver's hyperparameters within the PSA framework and to derive key principles for efficiently utilizing a limited time budget. We organize our findings into three main stages: for each solver, starting with Choco and then proceeding to ACE, first, we provide an aggregate analysis of the performance of PSA's internal component strategies, examining general trends across all problem instances. Second, based on these aggregate insights, we synthesize and present the champion configurations. Third, we then conduct detailed solver-specific performance comparisons, evaluating how these identified champion configurations perform against baseline configurations. For clarity and focus, the numerical results and pie charts presented in this section derive from a single, representative random seed.

We present the aggregated results in four comprehensive tables. Table~\ref{tab:choco_psa_strategies_combined} shows the frequency of winning PSA component strategies for the Choco solver with both BO and Hamming HPO methods. Table~\ref{tab:ace_psa_strategies_combined} provides the equivalent analysis for ACE. Based on these frequencies, we derive champion configurations for each solver-HPO combination in Table~\ref{tab:all_champion_configs}. Finally, Table~\ref{tab:performance_summary} summarizes the performance comparisons between PSA-tuned and default configurations.

\begin{table}[!tp]
\centering
\caption{PSA Component Strategy Frequencies for Choco Solver}
\label{tab:choco_psa_strategies_combined}
\begin{tabular}{lccc}
\toprule
\textbf{Component} & \textbf{Strategy} & \textbf{BO Wins} & \textbf{Hamming Wins} \\
\midrule
Global Time & Percent & 53.40\% & 42.98\% \\
& Dynamic & 46.60\% & 57.02\% \\
\midrule
Round Timeout & Static & 76.70\% & 81.58\% \\
& First-Runtime & 23.30\% & 18.42\% \\
\midrule
Timeout Evolution & DynamicGeometric & 31.07\% & 28.07\% \\
& DynamicLuby & 31.07\% & 20.18\% \\
& Static & 37.86\% & 51.75\% \\
\midrule
Stop Condition & Timeout & 36.89\% & 38.6\% \\
& Stagnation & 35.92\% & 27.19\% \\
& FirstSolution & 27.18\% & 34.21\% \\
\bottomrule
\end{tabular}
\end{table}

\begin{table}[!tp]
\centering
\caption{PSA Component Strategy Frequencies for ACE Solver}
\label{tab:ace_psa_strategies_combined}
\begin{tabular}{lccc}
\toprule
\textbf{Component} & \textbf{Strategy} & \textbf{BO Wins} & \textbf{Hamming Wins} \\
\midrule
Global Time & Percent & 50.44\% & 48.25\% \\
& Dynamic & 49.56\% & 51.75\% \\
\midrule
Round Timeout & Static & 77.18\% & 86.84\% \\
& First-Runtime & 22.82\% & 13.16\% \\
\midrule
Timeout Evolution & DynamicGeometric & 48.73\% & 27.19\% \\
& DynamicLuby & 25.17\% & 49.12\% \\
& Static & 26.10\% & 23.68\% \\
\midrule
Stop Condition & Timeout & 35.59\% & 29.82\% \\
& Stagnation & 33.94\% & 35.96\% \\
& FirstSolution & 30.46\% & 34.21\% \\
\bottomrule
\end{tabular}
\end{table}

\subsection{Choco Solver Results}

\paragraph{Analysis of PSA Component Strategies for Choco}
Table~\ref{tab:choco_psa_strategies_combined} reveals distinct patterns in effective PSA strategies for the Choco solver. For global time management, BO favors the \textit{percent} strategy (53.40\%), while Hamming prefers \textit{dynamic} allocation (57.02\%). Both HPO methods strongly favor \textit{static} round timeout initialization (76.70\% for BO, 81.58\% for Hamming). Timeout evolution patterns differ significantly: with BO, all three strategies are nearly equally effective (37.86\% static, 31.07\% each for dynamic variants), whereas with Hamming, \textit{static} evolution dominates (51.75\%). For stop conditions, \textit{timeout} is most frequent with both methods, though Hamming shows stronger preference for \textit{firstSolution} (34.21\%) compared to BO (27.18\%).

\paragraph{Synthesizing Champion Configurations for Choco}
Based on the aggregated analysis in Table~\ref{tab:choco_psa_strategies_combined}, we derive two champion configurations for Choco, one optimized for BO and another for Hamming. These configurations balance component effectiveness with overall strategy coherence and are summarized in Table~\ref{tab:all_champion_configs}. For BO, the champion combines \textit{percent} global time, \textit{static} timeout initialization, \textit{dynamicGeometric} evolution, and \textit{timeout} stop condition. For Hamming, the configuration uses \textit{dynamic} global time, \textit{static} timeout initialization, \textit{static} evolution, and \textit{timeout} stop condition.

\begin{table}[!tp]
\centering
\caption{Champion Configurations for Both Solvers}
\label{tab:all_champion_configs}

\setlength{\tabcolsep}{4pt}
\small
\begin{tabularx}{\linewidth}{@{}lXX@{}}
\toprule
\textbf{Component} &
\makecell{\textbf{Choco Champion}\\\textbf{(BO / Hamming)}} &
\makecell{\textbf{ACE Champion}\\\textbf{(BO / Hamming)}} \\
\midrule
Global Time Management & Percent / Dynamic & Percent / Dynamic \\
Round Timeout Initialization & Static / Static & Static / Static \\
Timeout Evolution Pattern & Dynamic\-Geometric / Static & Dynamic\-Geometric / Dynamic\-Luby \\
Probing Stop Condition & Timeout / Timeout & Timeout / Stagnation \\
\bottomrule
\end{tabularx}
\end{table}

\begin{figure}[!t]
\centering
\subfloat[PSA-BO Choco vs. Default Choco]{\label{fig:choco_pie_psa_bo_vs_default}%
\includegraphics[width=0.48\textwidth]{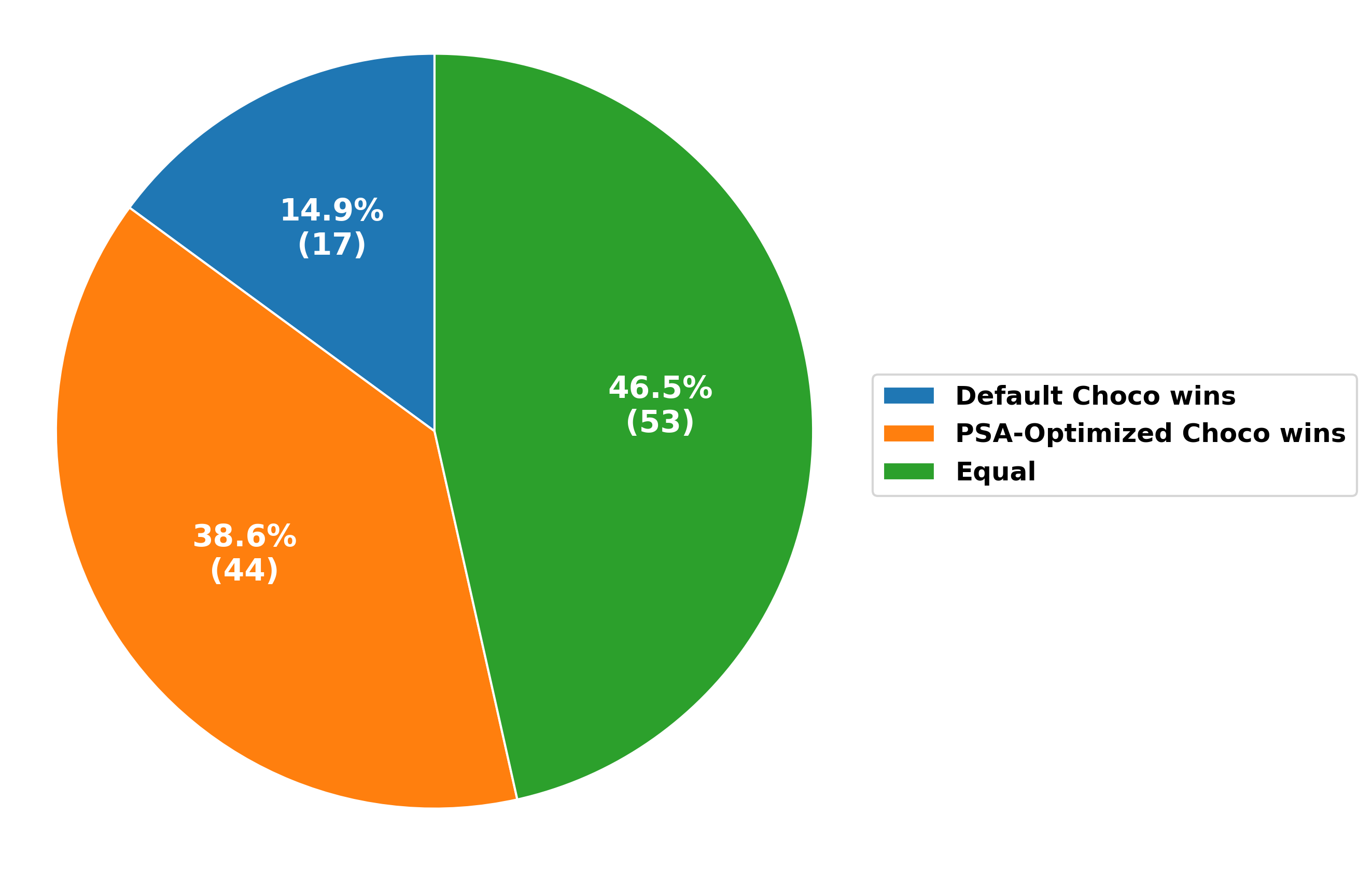}}%
\hfill
\subfloat[PSA-Hamming Choco vs. Default Choco]{\label{fig:choco_pie_psa_hamming_vs_default}%
\includegraphics[width=0.48\textwidth]{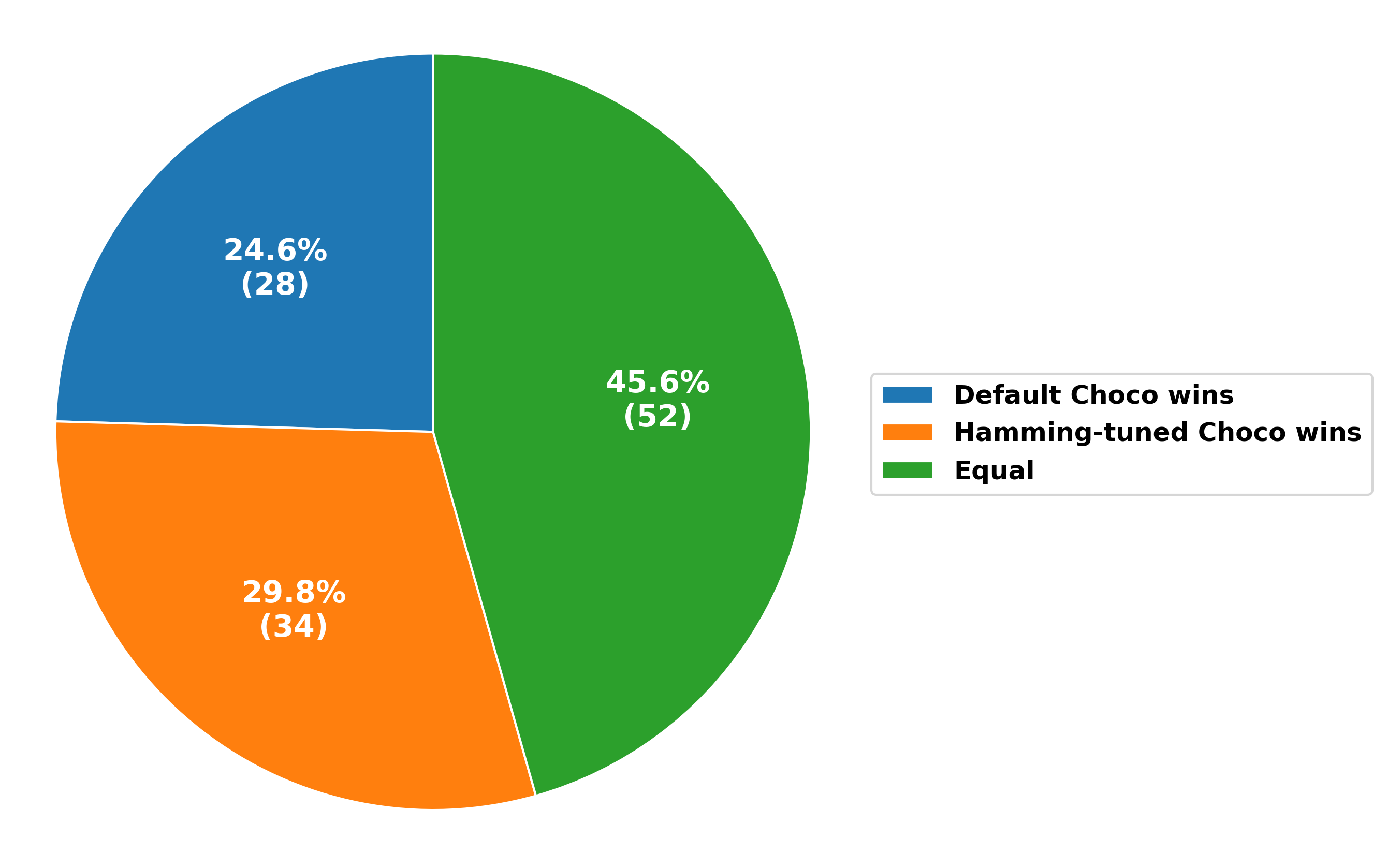}}%
\
\subfloat[PSA-BO Choco vs. PSA-Hamming Choco]{\label{fig:choco_pie_psa_bo_vs_hamming}%
\includegraphics[width=0.48\textwidth]{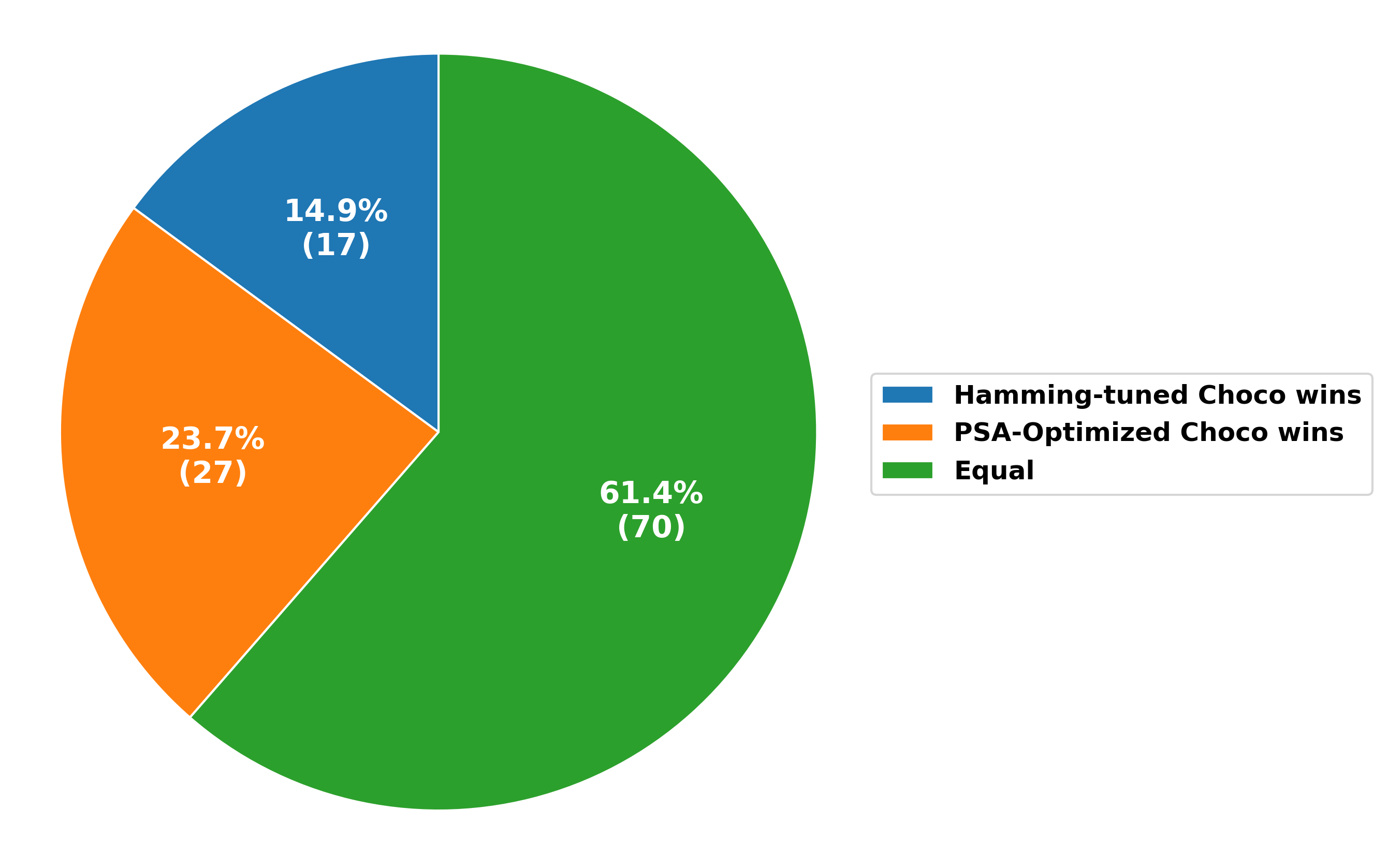}}%
\caption{Pairwise Performance Comparison for Choco Solver Approaches. (a) \textit{PSA-BO Choco} versus \textit{default Choco}. (b) \textit{PSA-Hamming Choco} versus \textit{default Choco}. (c) \textit{PSA-BO Choco} versus \textit{PSA-Hamming Choco}.}
\label{fig:choco_pie_comparisons_combined}
\end{figure}

\paragraph{Choco Performance Comparisons}
To evaluate PSA's effectiveness for Choco, we compare three approaches: default settings (\textit{default Choco}), PSA with BO (\textit{PSA-BO Choco}), and PSA with Hamming (\textit{PSA-Hamming Choco}). Results are summarized in Table~\ref{tab:performance_summary} and visualized in Figure~\ref{fig:choco_pie_comparisons_combined}.

\textit{PSA-BO Choco vs. Default Choco:} PSA-BO Choco performs better on 38.6\% of instances, while default Choco is superior on only 14.9\%, with ties on 46.5\% (Figure~\ref{fig:choco_pie_psa_bo_vs_default}). This demonstrates BO's advantage over default settings.

\textit{PSA-Hamming Choco vs. Default Choco:} Default Choco outperforms PSA-Hamming on 24.6\% of instances, while PSA-Hamming wins on 29.8\%, with ties on 45.6\% (Figure~\ref{fig:choco_pie_psa_hamming_vs_default}). This indicates PSA Hamming-based tuning generally outperforms Choco's default configurations.

\textit{PSA-BO Choco vs. PSA-Hamming Choco:} PSA-BO clearly outperforms PSA-Hamming, winning on 23.7\% of instances versus only 14.9\% for Hamming, with ties on 61.4\% (Figure~\ref{fig:choco_pie_psa_bo_vs_hamming}). This validates BO's superiority over the simpler Hamming approach.

\subsection{ACE Solver Results}

\paragraph{Analysis of PSA Component Strategies for ACE}
Table~\ref{tab:ace_psa_strategies_combined} shows PSA strategy effectiveness for ACE. Global time management shows minimal preference: BO slightly favors \textit{percent} (50.44\%), Hamming slightly favors \textit{dynamic} (51.75\%). Both strongly prefer \textit{static} timeout initialization (77.18\% BO, 86.84\% Hamming). Timeout evolution differs dramatically: BO strongly prefers \textit{dynamicGeometric} (48.73\%), while Hamming favors \textit{dynamicLuby} (49.12\%). Stop conditions show balanced distributions, with \textit{timeout} most frequent for BO (35.59\%) and \textit{stagnation} for Hamming (35.96\%).

\paragraph{Synthesizing Champion Configurations for ACE}
Based on Table~\ref{tab:ace_psa_strategies_combined}, we derive champion configurations for ACE (Table~\ref{tab:all_champion_configs}). For BO: \textit{percent} global time, \textit{static} timeout initialization, \textit{dynamicGeometric} evolution, and \textit{timeout} stop condition. For Hamming: \textit{dynamic} global time, \textit{static} timeout initialization, \textit{dynamicLuby} evolution, and \textit{stagnation} stop condition.

\begin{figure}[!t]
\centering
\subfloat[PSA-BO ACE vs. Default ACE]{\label{fig:ace_pie_psa_bo_vs_default}%
\includegraphics[width=0.48\textwidth]{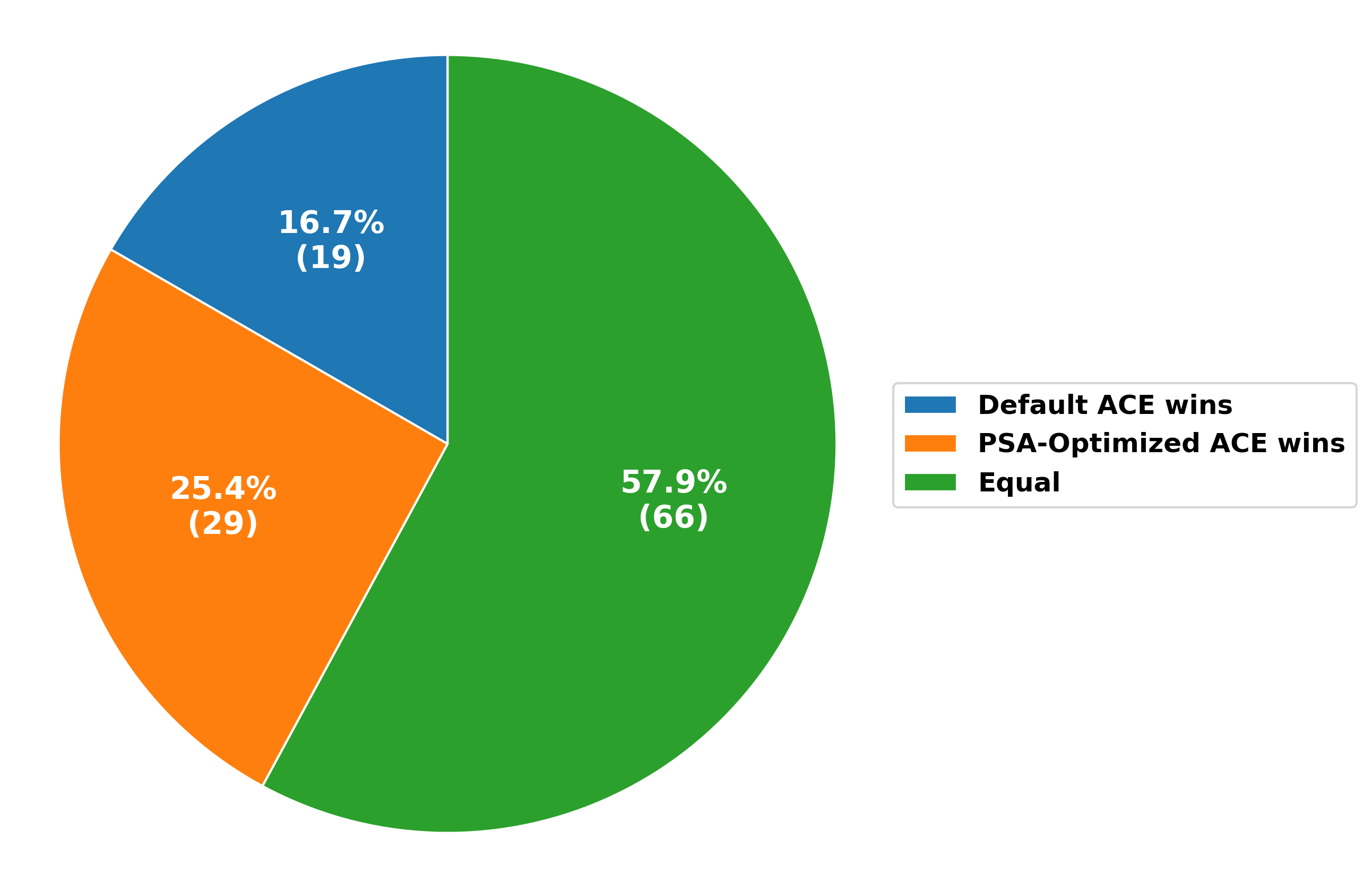}}%
\hfill
\subfloat[PSA-Hamming ACE vs. Default ACE]{\label{fig:ace_pie_psa_hamming_vs_default}%
\includegraphics[width=0.48\textwidth]{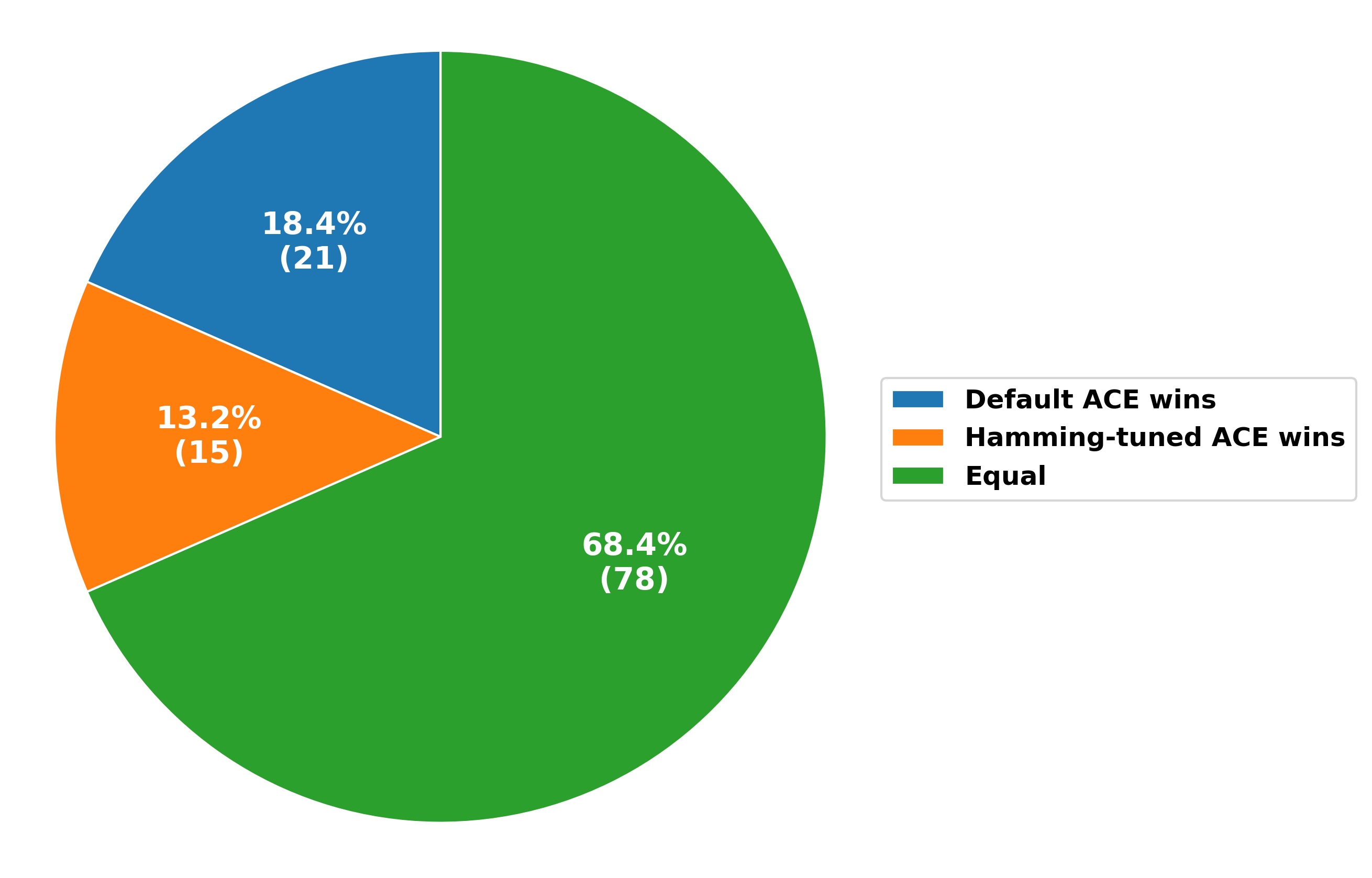}}%
\
\subfloat[PSA-BO ACE vs. PSA-Hamming ACE]{\label{fig:ace_pie_psa_bo_vs_hamming}%
\includegraphics[width=0.48\textwidth]{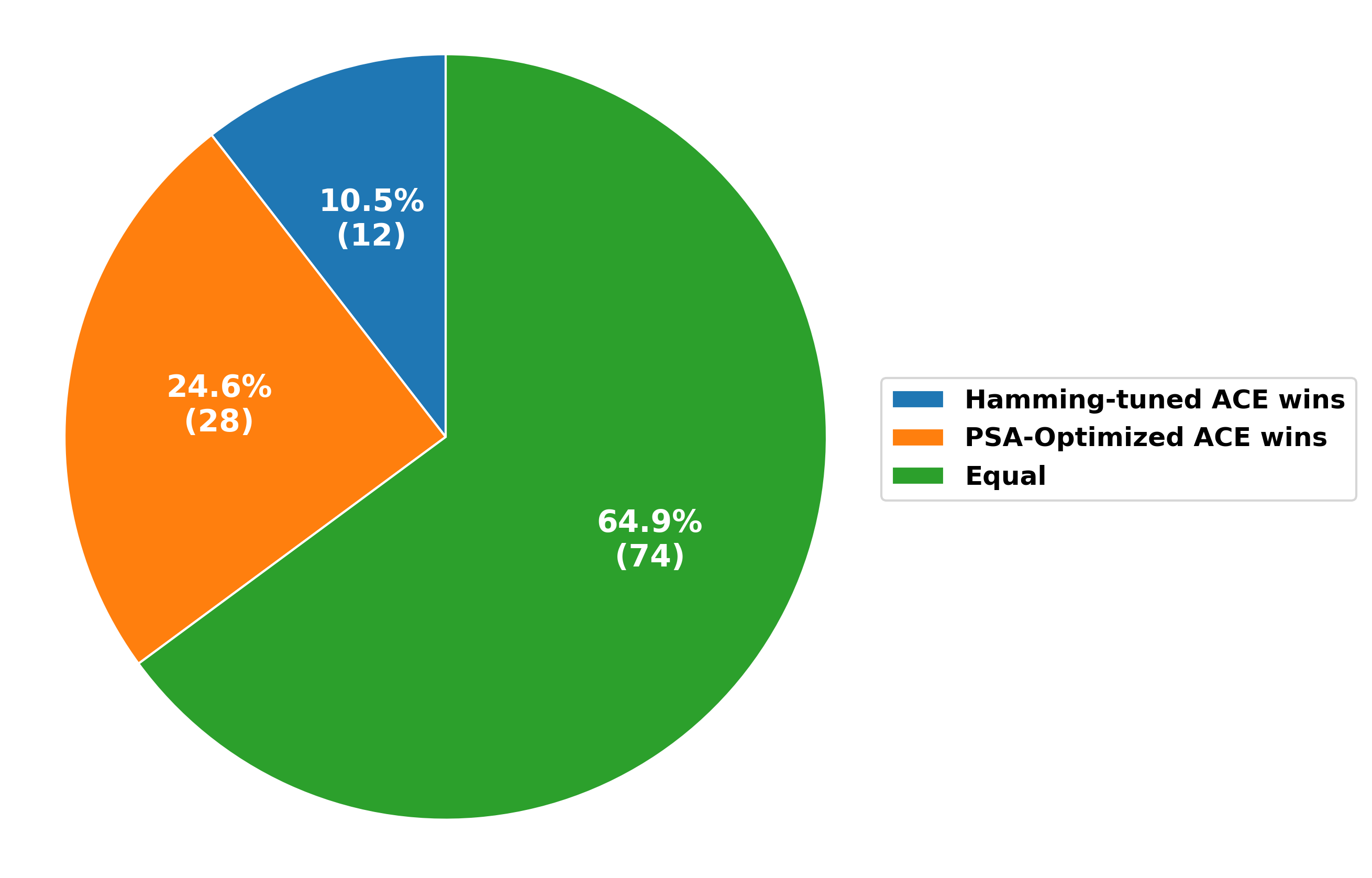}}%
\caption{Pairwise Performance Comparison for ACE Solver Approaches. (a) \textit{PSA-BO ACE} versus \textit{default ACE}. (b) \textit{PSA-Hamming ACE} versus \textit{default ACE}. (c) \textit{PSA-BO ACE} versus \textit{PSA-Hamming ACE}.}
\label{fig:ace_pie_comparisons_combined}
\end{figure}

\paragraph{Performance Comparison of ACE Configurations}
We compare default ACE with PSA-BO ACE and PSA-Hamming ACE. Results are in Table~\ref{tab:performance_summary} and Figure~\ref{fig:ace_pie_comparisons_combined}.

\begin{table}[!tp]
\centering
\caption{Performance Comparison Summary (Percentage of Instances)}
\label{tab:performance_summary}
\begin{tabular}{lccc}
\toprule
\textbf{Comparison} & \textbf{Winner A} & \textbf{Tie} & \textbf{Winner B} \\
\midrule
\textit{Choco Results} \\
PSA-BO Choco vs Default & 38.6\% & 46.5\% & 14.9\% \\
PSA-Hamming Choco vs Default & 29.8\% & 45.6\% & 24.6\% \\
PSA-BO vs PSA-Hamming Choco & 23.7\% & 61.4\% & 14.9\% \\
\midrule
\textit{ACE Results} \\
PSA-BO ACE vs Default & 25.4\% & 57.9\% & 16.7\% \\
PSA-Hamming ACE vs Default & 13.2\% & 68.4\% & 18.4\% \\
PSA-BO vs PSA-Hamming ACE & 24.6\% & 64.9\% & 10.5\% \\
\bottomrule
\end{tabular}
\end{table}

\textit{PSA-BO ACE vs. Default ACE:} PSA-BO ACE wins on 25.4\% of instances versus 16.7\% for default, with ties on 57.9\% (Figure~\ref{fig:ace_pie_psa_bo_vs_default}). BO provides clear advantage over default settings.

\textit{PSA-Hamming ACE vs. Default ACE:} Default ACE outperforms PSA-Hamming on 18.4\% of instances, while PSA-Hamming wins on 13.2\%, with ties on 68.4\% (Figure~\ref{fig:ace_pie_psa_hamming_vs_default}). Hamming-based tuning generally underperforms defaults.

\textit{PSA-BO ACE vs. PSA-Hamming ACE:} PSA-BO dominates, winning on 24.6\% of instances versus 10.5\% for Hamming, with ties on 64.9\% (Figure~\ref{fig:ace_pie_psa_bo_vs_hamming}). BO's model-based approach proves significantly more effective.

\section{Discussion and Future Research}
\label{sec:discussion}

Our experiments provide strong evidence that PSA is an effective strategy for optimizing constraint solvers under a fixed time budget. By utilizing BO to intelligently manage computational resources, PSA consistently identifies configurations that outperform baseline methods. This section interprets these findings, discusses their practical implications for users, and outlines promising avenues for future research.

The empirical results highlight several key trends. First, PSA demonstrates a clear advantage over the Hamming distance baseline. This validates the use of a model-based approach; the computational overhead of building a surrogate model in BO is justified by its ability to navigate complex search spaces far more efficiently than simple local search.

Furthermore, our analysis of the most successful configurations reveals that they typically combine two synergistic elements:
\begin{enumerate}
\item \textbf{Static Initialization:} A short, fixed time limit for the initial run acts as a gatekeeper, rapidly filtering out poor configurations without wasting significant time.
\item \textbf{Adaptive Exploration:} Following the initial filter, a dynamic evolution strategy increases the timeout. This ensures that if a configuration shows promise, the solver is granted sufficient time to deepen the search.
\end{enumerate}
This combination effectively balances the exploration-exploitation trade-off, minimizing risk while maximizing the probability of finding a high-quality solution. Consequently, the champion configurations we identified serve as robust generalists. While they may not be the theoretical optimum for every single instance, they provide a reliable, high-performance baseline across a wide variety of problem types. This robustness is particularly valuable for users who cannot afford the computational cost of tuning a solver separately for every new problem encounter.

These findings translate into three practical benefits for the CP community:
\begin{itemize}
\item \textbf{Democratization of Solving:} PSA automates the complex task of hyperparameter tuning. This pushes CP closer to the ideal model and solve paradigm, allowing users to focus on defining their problems rather than mastering the intricacies of solver configuration.
\item \textbf{Standardized Benchmarking:} The framework offers a principled methodology for comparing solvers under fixed time budgets, providing a scientific alternative to ad-hoc manual tuning.
\item \textbf{Solver Agnosticism:} The modular nature of PSA allows it to be easily applied to other solvers or algorithms, as demonstrated by its successful deployment on both ACE and Choco.
\end{itemize}

Despite these successes, there remain opportunities for further development. First, while our study focused on ACE and Choco, future work should validate the generalizability of PSA on other major solvers, such as OR-Tools. Second, the current implementation uses fixed meta-parameters (e.g., a 20\% probing budget). A promising direction is the development of a \textit{self-tuning} system that automatically adapts these parameters based on the problem size or complexity. Finally, rather than relying on a single static champion, future iterations could dynamically select a strategy class—switching between \textit{static} and \textit{dynamic} timeout evolution—based on the specific characteristics of the input problem.

\section{Conclusion}
\label{sec:conclusion}

This paper introduced PSA, a resource-aware, two-phase framework for automated HPO of CP solvers. PSA allocates a portion of the overall time budget to a probing phase, where BO guides a fast exploration of the configuration space, and uses the remaining time to solve the problem with the most promising configuration. The framework is fully integrated into CPMpy and includes the first integration of the ACE solver into the library.

Our evaluation across 114 benchmark instances shows that PSA provides consistent benefits for two solvers with fundamentally different architectures. For ACE, PSA with BO provided clear benefits over the default configuration: it achieved better results in 25.4\% of the instances (29 cases), while the default configuration was better in only 16.7\% (19 cases); both performed equally on 57.9\% (66 cases). For Choco, PSA with BO improved performance in 38.6\% of instances and matched the baseline in 46.5\%, while the default configuration was better in only 14.9\% of cases.

Across both solvers, PSA with BO consistently outperformed the Hamming distance baseline. For ACE, PSA-BO won 24.6\% of comparisons against PSA-Hamming, while PSA-Hamming won only 10.5\%. For Choco, PSA-BO won 23.7\% of comparisons against PSA-Hamming, while PSA-Hamming won 14.9\%. These results demonstrate the advantage of model-guided BO over simpler local-perturbation strategies.

Beyond numerical improvements, the results highlight an important insight: the effectiveness of a tuning strategy depends strongly on the type of problem. Harder optimization problems benefit from dynamic timeout evolution, which allows PSA to progressively allocate more time to promising configurations, while easier satisfaction-oriented instances benefit from evaluating many configurations quickly under static time budgets. This indicates that PSA not only improves solver performance but also exposes useful structure in how hyperparameters interact with different classes of problems.

Overall, PSA contributes a practical and automated approach for improving the performance of constraint solvers without requiring expert knowledge or manual parameter tuning. Future work will focus on making PSA more adaptive by incorporating instance features, supporting multi-objective configuration, and extending the framework toward general automated solver design.

\section*{Acknowledgment}

\subsection*{Funding}
This work is partially funded by the joint research program UL/SnT—ILNAS on Technical Standardization for Trustworthy ICT, Aerospace, and Construction, and supported by the Luxembourg National Research Fund (FNR)—COMOC Project, ref. C21/IS/16101289.

\subsection*{Data Availability}
The benchmark instances used in this study are publicly available as part of the XCSP24 benchmark suite at: 
\url{https://www.cril.univ-artois.fr/XCSP24/}.

\subsection*{Code Availability} 
The implementation of the Probe and Solve Algorithm (PSA) integrated into CPMpy is available at:
\url{https://github.com/Hedieh-Haddad/cpmpy/tree/psa}.


\backmatter






\bibliography{references}

@misc{cpmpy_cpmpy_2025,
	title = {{CPMpy}: {Constraint} {Programming} and {Modeling} in {Python} — {CPMpy} 0.9.24 documentation},
	url = {https://cpmpy.readthedocs.io/en/latest/},
	urldate = {2025-10-16},
	author = {CPMpy, Development Team},
	year = {2025},
}

@book{regin_generalized_1996,
	address = {Portland, Oregon, USA},
	title = {Generalized {Arc} {Consistency} for {Global} {Cardinality} {Constraint}.},
	isbn = {0-262-51091-X},
	publisher = {AAAI Press},
	author = {Régin, Jean-Charles},
	month = jan,
	year = {1996},
	note = {Journal Abbreviation: Proceedings AAAI'96
Pages: 215
Publication Title: Proceedings AAAI'96},
}

@inproceedings{pesant_view_1996,
	address = {Cambridge, US},
	title = {View of local search in constraint programming},
	volume = {1118},
	isbn = {978-3-540-61551-4},
	url = {https://doi.org/10.1007/3-540-61551-2},
	doi = {10.1007/3-540-61551-2},
	language = {en},
	urldate = {2025-10-15},
	booktitle = {Principles and {Practice} of {Constraint} {Programming} - {CP}'96},
	publisher = {Springer-Verlag},
	author = {Pesant, Gilles and Gendreau, Michel},
	year = {1996},
	pages = {353--366},
}

@misc{prudhomme_choco-solver_2022,
	title = {Choco-solver: {A} {Java} library for constraint programming},
	copyright = {BSD-4-Clause},
	shorttitle = {Choco-solver},
	url = {https://github.com/chocoteam/choco-solver},
	doi = {10.21105/joss.04708},
	urldate = {2024-06-27},
	author = {Prud'homme, Charles and Fages, Jean-Guillaume},
	year = {2022},
	note = {Issue: 78
Pages: 4708
Publication Title: Journal of Open Source Software
Volume: 7
original-date: 2011-11-04},
}

@inproceedings{bergstra_algorithms_2011,
	address = {Cambridge, US},
	title = {Algorithms for {Hyper}-{Parameter} {Optimization}},
	volume = {24},
	url = {https://proceedings.neurips.cc/paper-files/paper/2011/hash/86e8f7ab32cfd12577bc2619bc635690-Abstract.html},
	urldate = {2025-03-28},
	booktitle = {Advances in {Neural} {Information} {Processing} {Systems}},
	publisher = {Curran Associates, Inc.},
	author = {Bergstra, James and Bardenet, Rémi and Bengio, Yoshua and Kégl, Balázs},
	year = {2011},
}

@article{wu_hyperparameter_2019,
	title = {Hyperparameter {Optimization} for {Machine} {Learning} {Models} {Based} on {Bayesian} {Optimization}},
	volume = {17},
	issn = {1674-862X},
	url = {https://www.sciencedirect.com/science/article/pii/S1674862X19300047},
	doi = {10.11989/JEST.1674-862X.80904120},
	number = {1},
	urldate = {2025-04-16},
	journal = {Journal of Electronic Science and Technology},
	author = {Wu, Jia and Chen, Xiu-Yun and Zhang, Hao and Xiong, Li-Dong and Lei, Hang and Deng, Si-Hao},
	month = mar,
	year = {2019},
	pages = {26--40},
}

@incollection{amini_learning_2023,
	address = {Cham},
	title = {Learning {Optimal} {Decision} {Trees} {Under} {Memory} {Constraints}},
	volume = {13717},
	isbn = {978-3-031-26418-4 978-3-031-26419-1},
	url = {https://link.springer.com/10.1007/978-3-031-26419-1},
	doi = {10.1007/978-3-031-26419-1},
	language = {en},
	urldate = {2026-01-04},
	booktitle = {Machine {Learning} and {Knowledge} {Discovery} in {Databases}},
	publisher = {Springer Nature Switzerland},
	author = {Aglin, Gaël and Nijssen, Siegfried and Schaus, Pierre},
	editor = {Amini, Massih-Reza and Canu, Stéphane and Fischer, Asja and Guns, Tias and Kralj Novak, Petra and Tsoumakas, Grigorios},
	year = {2023},
	note = {Series Title: Lecture Notes in Computer Science},
	pages = {393--409},
}

@article{osullivan_opportunities_2021,
	title = {Opportunities and {Challenges} for {Constraint} {Programming}},
	volume = {26},
	issn = {2374-3468, 2159-5399},
	url = {https://ojs.aaai.org/index.php/AAAI/article/view/8450},
	doi = {10.1609/aaai.v26i1.8450},
	language = {en},
	number = {1},
	urldate = {2026-01-04},
	journal = {Proceedings of the AAAI Conference on Artificial Intelligence},
	author = {O'Sullivan, Barry},
	month = sep,
	year = {2021},
	pages = {2148--2152},
}

@book{baptiste_constraint-based_2001,
	address = {Amsterdam},
	title = {Constraint-{Based} {Scheduling}: {Applying} {Constraint} {Programming} to {Scheduling} {Problems}},
	isbn = {978-0-7923-7408-4},
	shorttitle = {Constraint-{Based} {Scheduling}},
	language = {en},
	publisher = {Springer Science \& Business Media},
	author = {Baptiste, Philippe and Pape, Claude Le and Nuijten, Wim},
	month = jul,
	year = {2001},
}

@inproceedings{bessiere_complexity_2004,
	address = {San Jose, California},
	series = {{AAAI}'04},
	title = {The complexity of global constraints},
	isbn = {978-0-262-51183-4},
	urldate = {2026-01-01},
	booktitle = {Proceedings of the 19th national conference on {Artifical} intelligence},
	publisher = {AAAI Press},
	author = {Bessiere, Christian and Hebrard, Emmanuel and Hnich, Brahim and Walsh, Toby},
	month = jul,
	year = {2004},
	pages = {112--117},
}

@incollection{schaus_model-based_2022,
	address = {Cham},
	title = {Model-{Based} {Algorithm} {Configuration} with {Adaptive} {Capping} and {Prior} {Distributions}},
	volume = {13292},
	isbn = {978-3-031-08010-4},
	url = {https://link.springer.com/10.1007/978-3-031-08011-1},
	doi = {10.1007/978-3-031-08011-1},
	language = {en},
	urldate = {2025-03-28},
	booktitle = {Integration of {Constraint} {Programming}, {Artificial} {Intelligence}, and {Operations} {Research}},
	publisher = {Springer International Publishing},
	author = {Bleukx, Ignace and Berden, Senne and Coenen, Lize and Decleyre, Nicholas and Guns, Tias},
	editor = {Schaus, Pierre},
	year = {2022},
	note = {Series Title: Lecture Notes in Computer Science},
	pages = {64--73},
}

@incollection{bessiere_chapter_2006,
	address = {Amsterdam},
	series = {Handbook of {Constraint} {Programming}},
	title = {Chapter 3 - {Constraint} {Propagation}},
	volume = {2},
	url = {https://www.sciencedirect.com/science/article/pii/S1574652606800076},
	doi = {10.1016/S1574-6526(06)80007-6},
	urldate = {2025-10-15},
	booktitle = {Foundations of {Artificial} {Intelligence}},
	publisher = {Elsevier},
	author = {Bessiere, Christian},
	editor = {Rossi, Francesca and van Beek, Peter and Walsh, Toby},
	month = jan,
	year = {2006},
	pages = {29--83},
}

@article{cacchiani_knapsack_2022,
	title = {Knapsack problems — {An} overview of recent advances. {Part} {II}: {Multiple}, multidimensional, and quadratic knapsack problems},
	volume = {143},
	issn = {0305-0548},
	shorttitle = {Knapsack problems — {An} overview of recent advances. {Part} {II}},
	url = {https://www.sciencedirect.com/science/article/pii/S0305054821003889},
	doi = {10.1016/j.cor.2021.105693},
	urldate = {2026-01-01},
	journal = {Computers \& Operations Research},
	author = {Cacchiani, Valentina and Iori, Manuel and Locatelli, Alberto and Martello, Silvano},
	month = jul,
	year = {2022},
	pages = {105693},
}

@incollection{feurer_hyperparameter_2019,
	address = {Cham},
	series = {The {Springer} {Series} on {Challenges} in {Machine} {Learning}},
	title = {Hyperparameter {Optimization}},
	isbn = {978-3-030-05318-5},
	url = {https://doi.org/10.1007/978-3-030-05318-5},
	doi = {10.1007/978-3-030-05318-5_1},
	language = {en},
	urldate = {2024-01-13},
	booktitle = {Automated {Machine} {Learning}: {Methods}, {Systems}, {Challenges}},
	publisher = {Springer International Publishing},
	author = {Feurer, Matthias and Hutter, Frank},
	editor = {Hutter, Frank and Kotthoff, Lars and Vanschoren, Joaquin},
	year = {2019},
	pages = {3--33},
}

@incollection{rasmussen_gaussian_2004,
	address = {Berlin, Heidelberg},
	title = {Gaussian {Processes} in {Machine} {Learning}},
	isbn = {978-3-540-28650-9},
	url = {https://doi.org/10.1007/978-3-540-28650-9},
	doi = {10.1007/978-3-540-28650-9_4},
	language = {en},
	urldate = {2025-05-20},
	booktitle = {Advanced {Lectures} on {Machine} {Learning}: {ML} {Summer} {Schools} 2003, {Canberra}, {Australia}, {February} 2 - 14, 2003, {Tübingen}, {Germany}, {August} 4 - 16, 2003, {Revised} {Lectures}},
	publisher = {Springer},
	author = {Rasmussen, Carl Edward},
	editor = {Bousquet, Olivier and von Luxburg, Ulrike and Rätsch, Gunnar},
	year = {2004},
	pages = {63--71},
}

@article{bergstra_random_2012,
	title = {Random search for hyper-parameter optimization},
	volume = {13},
	issn = {1532-4435},
	number = {null},
	journal = {J. Mach. Learn. Res.},
	author = {Bergstra, James and Bengio, Yoshua},
	month = feb,
	year = {2012},
	pages = {281--305},
}

@book{hutter_automated_2019,
	address = {Cham},
	series = {The {Springer} {Series} on {Challenges} in {Machine} {Learning}},
	title = {Automated {Machine} {Learning}: {Methods}, {Systems}, {Challenges}},
	copyright = {https://creativecommons.org/licenses/by/4.0},
	isbn = {978-3-030-05317-8},
	shorttitle = {Automated {Machine} {Learning}},
	url = {http://link.springer.com/10.1007/978-3-030-05318-5},
	doi = {10.1007/978-3-030-05318-5},
	language = {en},
	urldate = {2025-10-15},
	publisher = {Springer International Publishing},
	editor = {Hutter, Frank and Kotthoff, Lars and Vanschoren, Joaquin},
	year = {2019},
}

@book{lecoutre_constraint_2009,
	address = {Hoboken, NJ},
	title = {Constraint networks: techniques and algorithms},
	isbn = {978-1-84821-106-3},
	shorttitle = {Constraint networks},
	doi = {10.1002/9780470611821},
	language = {en},
	publisher = {Wiley},
	author = {Lecoutre, Christophe},
	year = {2009},
}

@inproceedings{haddad_selecting_2024,
	title = {Selecting {Search} {Strategy} in {Constraint} {Solvers} using {Bayesian} {Optimization}},
	issn = {2375-0197},
	url = {https://ieeexplore.ieee.org/abstract/document/10849432},
	doi = {10.1109/ICTAI62512.2024.00113},
	urldate = {2025-08-01},
	booktitle = {2024 {IEEE} 36th {International} {Conference} on {Tools} with {Artificial} {Intelligence} ({ICTAI})},
	author = {Haddad, Hedieh and Talbot, Pierre and Bouvry, Pascal},
	month = oct,
	year = {2024},
	pages = {764--773},
}

@misc{bischl_hyperparameter_2021,
	title = {Hyperparameter {Optimization}: {Foundations}, {Algorithms}, {Best} {Practices} and {Open} {Challenges}},
	shorttitle = {Hyperparameter {Optimization}},
	url = {http://arxiv.org/abs/2107.05847},
	doi = {10.48550/arXiv.2107.05847},
	urldate = {2025-05-20},
	publisher = {arXiv},
	author = {Bischl, Bernd and Binder, Martin and Lang, Michel and Pielok, Tobias and Richter, Jakob and Coors, Stefan and Thomas, Janek and Ullmann, Theresa and Becker, Marc and Boulesteix, Anne-Laure and Deng, Difan and Lindauer, Marius},
	month = nov,
	year = {2021},
	note = {arXiv:2107.05847 [stat]},
}

@article{kushner_new_1964,
	title = {A {New} {Method} of {Locating} the {Maximum} {Point} of an {Arbitrary} {Multipeak} {Curve} in the {Presence} of {Noise}},
	volume = {86},
	issn = {0021-9223},
	url = {https://asmedigitalcollection.asme.org/fluidsengineering/article/86/1/97/392213/A-New-Method-of-Locating-the-Maximum-Point-of-an},
	doi = {10.1115/1.3653121},
	language = {en},
	number = {1},
	urldate = {2025-05-20},
	journal = {Journal of Basic Engineering},
	author = {Kushner, H. J.},
	month = mar,
	year = {1964},
	pages = {97--106},
}

@inproceedings{gan_acquisition_2021,
	title = {Acquisition {Functions} in {Bayesian} {Optimization}},
	url = {https://ieeexplore.ieee.org/document/9696089},
	doi = {10.1109/ICBASE53849.2021.00032},
	urldate = {2025-05-20},
	booktitle = {2021 2nd {International} {Conference} on {Big} {Data} \& {Artificial} {Intelligence} \& {Software} {Engineering} ({ICBASE})},
	author = {Gan, Weiao and Ji, Ziyuan and Liang, Yongqing},
	month = sep,
	year = {2021},
	pages = {129--135},
}

@article{amadini_portfolio_2016,
	title = {Portfolio approaches for constraint optimization problems},
	volume = {76},
	issn = {1573-7470},
	url = {https://doi.org/10.1007/s10472-015-9459-5},
	doi = {10.1007/s10472-015-9459-5},
	language = {en},
	number = {1},
	urldate = {2025-04-16},
	journal = {Annals of Mathematics and Artificial Intelligence},
	author = {Amadini, Roberto and Gabbrielli, Maurizio and Mauro, Jacopo},
	month = feb,
	year = {2016},
	pages = {229--246},
}

@misc{boussemart_xcsp3-core_2024,
	title = {{XCSP3}-core: {A} {Format} for {Representing} {Constraint} {Satisfaction}/{Optimization} {Problems}},
	shorttitle = {{XCSP3}-core},
	url = {http://arxiv.org/abs/2009.00514},
	doi = {10.48550/arXiv.2009.00514},
	urldate = {2025-04-16},
	publisher = {arXiv},
	author = {Boussemart, Frédéric and Lecoutre, Christophe and Audemard, Gilles and Piette, Cédric},
	month = aug,
	year = {2024},
	note = {arXiv:2009.00514 [cs]},
}

@misc{lecoutre_ace_2024,
	title = {{ACE}, a generic constraint solver},
	url = {http://arxiv.org/abs/2302.05405},
	doi = {10.48550/arXiv.2302.05405},
	urldate = {2025-04-16},
	publisher = {arXiv},
	author = {Lecoutre, Christophe},
	month = sep,
	year = {2024},
	note = {arXiv:2302.05405 [cs]},
}

@inproceedings{hutter_sequential_2011,
	address = {Berlin, Heidelberg},
	title = {Sequential {Model}-{Based} {Optimization} for {General} {Algorithm} {Configuration}},
	isbn = {978-3-642-25566-3},
	doi = {10.1007/978-3-642-25566-3_40},
	language = {en},
	booktitle = {Learning and {Intelligent} {Optimization}},
	publisher = {Springer},
	author = {Hutter, Frank and Hoos, Holger H. and Leyton-Brown, Kevin},
	editor = {Coello, Carlos A. Coello},
	year = {2011},
	pages = {507--523},
}

@article{hutter_paramils_2009,
	title = {{ParamILS}: {An} {Automatic} {Algorithm} {Configuration} {Framework}},
	volume = {36},
	issn = {1076-9757},
	shorttitle = {{ParamILS}},
	url = {http://arxiv.org/abs/1401.3492},
	doi = {10.1613/jair.2861},
	urldate = {2025-01-21},
	journal = {Journal of Artificial Intelligence Research},
	author = {Hutter, Frank and Stuetzle, Thomas and Leyton-Brown, Kevin and Hoos, Holger H.},
	month = oct,
	year = {2009},
	note = {arXiv:1401.3492 [cs]},
	pages = {267--306},
}

@book{apt_principles_2003,
	address = {Cambridge},
	title = {Principles of {Constraint} {Programming}},
	isbn = {978-0-521-82583-2},
	url = {https://www.cambridge.org/core/books/principles-of-constraint-programming/C008FB32571F66C3EE0EEEBDE1F98A7D},
	doi = {10.1017/CBO9780511615320},
	urldate = {2025-01-17},
	publisher = {Cambridge University Press},
	author = {Apt, Krzysztof},
	year = {2003},
}

@article{hooker_constraint_2018,
	title = {Constraint programming and operations research},
	volume = {23},
	issn = {1383-7133, 1572-9354},
	url = {http://link.springer.com/10.1007/s10601-017-9280-3},
	doi = {10.1007/s10601-017-9280-3},
	language = {en},
	number = {2},
	urldate = {2025-01-17},
	journal = {Constraints},
	author = {Hooker, J. N. and Van Hoeve, W.-J.},
	month = apr,
	year = {2018},
	pages = {172--195},
}

@article{marty_learning_2024,
	title = {Learning and fine-tuning a generic value-selection heuristic inside a constraint programming solver},
	issn = {1572-9354},
	url = {https://doi.org/10.1007/s10601-024-09377-4},
	doi = {10.1007/s10601-024-09377-4},
	language = {en},
	urldate = {2024-12-13},
	journal = {Constraints},
	author = {Marty, Tom and Boisvert, Léo and François, Tristan and Tessier, Pierre and Gautier, Louis and Rousseau, Louis-Martin and Cappart, Quentin},
	month = nov,
	year = {2024},
}

@inproceedings{hutter_automatic_2007,
	address = {Vancouver, British Columbia, Canada},
	series = {{AAAI}'07},
	title = {Automatic algorithm configuration based on local search},
	isbn = {978-1-57735-323-2},
	urldate = {2024-11-25},
	booktitle = {Proceedings of the 22nd national conference on {Artificial} intelligence - {Volume} 2},
	publisher = {AAAI Press},
	author = {Hutter, Frank and Hoos, Holger H. and Stützle, Thomas},
	month = jul,
	year = {2007},
	pages = {1152--1157},
}

@inproceedings{brause_combining_2022,
	address = {Vienna, Austria},
	title = {Combining {Constraint} {Solving} and {Bayesian} {Techniques} for {System} {Optimization}},
	isbn = {978-1-956792-00-3},
	url = {https://www.ijcai.org/proceedings/2022/249},
	doi = {10.24963/ijcai.2022/249},
	language = {en},
	urldate = {2024-11-25},
	booktitle = {Proceedings of the {Thirty}-{First} {International} {Joint} {Conference} on {Artificial} {Intelligence}},
	publisher = {International Joint Conferences on Artificial Intelligence Organization},
	author = {Brauße, Franz and Khasidashvili, Zurab and Korovin, Konstantin},
	month = jul,
	year = {2022},
	pages = {1788--1794},
}

@misc{rossi_handbook_2006,
	title = {Handbook of {Constraint} {Programming} [{Book}]},
	url = {https://www.oreilly.com/library/view/handbook-of-constraint/9780444527264/},
	language = {en},
	urldate = {2024-06-18},
	author = {Rossi, Francesca and Beek, Peter van and Walsh, Toby},
	month = aug,
	year = {2006},
	note = {ISBN: 9780080463803},
}

@misc{ungredda_bayesian_2021,
	title = {Bayesian {Optimisation} for {Constrained} {Problems}},
	url = {http://arxiv.org/abs/2105.13245},
	doi = {10.48550/arXiv.2105.13245},
	urldate = {2024-07-04},
	publisher = {arXiv},
	author = {Ungredda, Juan and Branke, Juergen},
	month = may,
	year = {2021},
	note = {arXiv:2105.13245 [cs, stat]},
}

@misc{candelieri_mastering_2023,
	title = {Mastering the exploration-exploitation trade-off in {Bayesian} {Optimization}},
	url = {http://arxiv.org/abs/2305.08624},
	doi = {10.48550/arXiv.2305.08624},
	urldate = {2024-06-27},
	publisher = {arXiv},
	author = {Candelieri, Antonio},
	month = may,
	year = {2023},
	note = {arXiv:2305.08624 [cs, math]},
}

@misc{liashchynskyi_grid_2019,
	title = {Grid {Search}, {Random} {Search}, {Genetic} {Algorithm}: {A} {Big} {Comparison} for {NAS}},
	shorttitle = {Grid {Search}, {Random} {Search}, {Genetic} {Algorithm}},
	url = {http://arxiv.org/abs/1912.06059},
	doi = {10.48550/arXiv.1912.06059},
	urldate = {2024-01-13},
	publisher = {arXiv},
	author = {Liashchynskyi, Petro and Liashchynskyi, Pavlo},
	month = dec,
	year = {2019},
	note = {arXiv:1912.06059 [cs, stat]},
}

\end{document}